# Deep Learning on 3D Semantic Segmentation: A Detailed Review

Thodoris Betsas[1*], Andreas Georgopoulos[1], Anastasios Doulamis[1], Pierre Grussenmeyer[2]

[1]National Technical University of Athens (NTUA), School of Rural Surveying and Geoinformatics Engineering, Lab. of Photogrammetry
[2]Université de Strasbourg, INSA Strasbourg, CNRS, ICube Laboratory UMR 7357, Photogrammetry and Geomatics Group, 67000

betsasth@mail.ntua.gr, drag@central.ntua.gr, adoulam@cs.ntua.gr, pierre.grussenmeyer@insa-strasbourg.fr

KEYWORDS: 3D Semantic Segmentation, Point Clouds, Deep Learning, Review

## ABSTRACT:

In this paper an exhaustive review and comprehensive analysis of recent and former deep learning methods in 3D Semantic Segmentation (3DSS) is presented. In the related literature, the taxonomy scheme used for the classification of the 3DSS deep learning methods is ambiguous. Based on the taxonomy schemes of 9 existing review papers, a new taxonomy scheme of the 3DSS deep learning methods is proposed, aiming to standardize it and improve the comparability and clarity across related studies. Furthermore, an extensive overview of the available 3DSS indoor and outdoor datasets is provided along with their links. The core part of the review is the detailed presentation of recent and former 3DSS deep learning methods and their classification using the proposed taxonomy scheme along with their GitHub repositories. Additionally, a brief but informative analysis of the evaluation metrics and loss functions used in 3DSS is included. Finally, a fruitful discussion of the examined 3DSS methods and datasets, is presented to foster new research directions and applications in the field of 3DSS. Supplementary, to this review a GitHub repository is provided (https://github.com/thobet/Deep-Learning-on-3D-Semantic-Segmentation-a-Detailed-Review) including a quick classification of over 400 3DSS methods, using the proposed taxonomy scheme.

## 1. INTRODUCTION

Acquiring 3D information from various environments of the world is becoming easier day to day, due to the wealth of the available sensors like the ALS, TLS, LiDAR, RADAR, DLSR and Sonar. Most of the available sensors produce 3D point clouds which, before the post processing steps e.g., filtering, constitute their raw 3D information. In general, the deep learning (DL) algorithms dominate several 2D tasks like classification, object detection, semantic -instance and -panoptic segmentation. Meanwhile, the 3D point clouds have unique benefits in comparison to the 2D images such as they capture the detailed geometry of the objects [1], [2] and intuitively they are similar to the 3D world. However, the application of 2D DL methods directly into 3D space using 3D point clouds, is not a straightforward process due to their properties.

In detail, the 3D point clouds are commonly unordered, unstructured and irregular. More concretely, unordered means that the 3D points are not on a regular grid, like the pixels of the images, unstructured means that the 3D points do not carry the neighboring points information and finally irregular means that some 3D point clouds contain regions with different point densities i.e., the 3D points are not evenly sampled across the different regions of the scene. Moreover, each point's neighborhood forms meaningful information about the characteristics of its region i.e., the 3D points are not isolated. Also, the 3D point clouds are invariant under transformations i.e., rotating and translating the 3D points all together do not alter the characteristics of the acquired scene. To conclude, consider that there is a terminology confusion among the meaning of the 3D point cloud properties along the 3D Semantic

Segmentation (3DSS) papers i.e., the meaning of the unordered, irregular and unstructured properties, however the actual properties remain the same.

In general, semantic segmentation (SS) is defined as the association of each element of the data under process with a meaningful label. In this regard, 2DSS using images aims to assign a meaningful label to each pixel of the image while 3DSS using point clouds aims to assign a meaningful label to each 3D point of the point cloud etc. In fact, there is a terminology confusion regarding the 3DSS. To be more specific, 3DSS is commonly referred to as 3D point cloud classification, 3D point cloud per point classification or 3D labeling. However, throughout this effort, the 3DSS terminology is preferred.

To sum up, the contributions of this effort varied. First and foremost, Section 2 describes in detail and compares the taxonomy schemes of the 3DSS DL methods proposed by previous 3DSS review papers, resulting safely to a unified taxonomy scheme. Additionally, the general idea behind the selection of such categories, based on the examined 3DSS methods, is thoroughly described. Apart from the general categories of the 3DSS DL methods, the same analysis is conducted for their subcategories. Moreover, in Section 3 over 40 3DSS indoor and outdoor datasets are presented including their links. In addition to the available datasets, the commonly used evaluation metrics of the 3DSS algorithms are presented in Section 4. Afterwards, a detailed analysis of the collected 3DSS algorithms of each category and subcategory defined in Section 2, is presented in Section 5. Notable, there are included subcategories also for the Hybrid based methods apart from the main categories i.e., -Point, -Dimensionality Reduction, -Discretization and Graph based methods. Then, a brief but informative analysis of the 3DSS loss functions is presented in Section 6. Finally, in Section 7 a fruitful analysis of the examined 3DSS algorithms and datasets, is presented to foster new research directions and applications in the field of 3DSS. To the best of our knowledge this effort is the first 3DSS review paper including a Taxonomy Scheme section i.e., Section 2. Finally, a GitHub repository (https://github.com/thobet/Deep-Learning-on-3D-Semantic-Segmentation-a-Detailed-Review) which includes a quick classification of over 400 3DSS algorithms is included. In addition to this effort, the GitHub repository offers a valuable source for the investigation of both traditional [3], [4], [5], [6] and deep learning 3DSS algorithms.

## 2. TAXONOMY SCHEME OF DEEP LEARNING 3D SEMANTIC SEGMENTATION METHODS

The last decade many researchers published review and benchmark papers [7], [8], [9], [10], [11], [12], [13], [14], [15], [16], [17], [18], [19], [20], [21], [21], [22] presenting the ongoing research on 3D Semantic Segmentation (3DSS) i.e., the open challenges and research questions, the difficulties presented in 3DSS applications, ideas for future work and many 3DSS algorithms. In this section, we present a literature review of 3DSS review papers and we aim to draw a safe conclusion about a taxonomy scheme of 3DSS deep learning methods. To be more specific, each review paper proposes a different categorization scheme of the existing 3DSS algorithms using different names for each category. But are those categories really different?

**Previous 3DSS Review Papers:** First and foremost, 3DSS algorithms could be divided into two broad categories, *Regular Supervised Machine Learning Methods* and *Deep Learning Methods* [18], [19], [22]. The methods belonging to the first category are often called traditional methods while the methods belong to the second one deep learning methods. The main difference between the two categories is the feature extraction approach. Traditional methods rely on handcrafted features while the deep learning approaches learn to extract the features from the data without user involvement. Traditional methods are used in many applications e.g., in Cultural Heritage (CH) domain [11] but in this section we focus on the deep learning methods for 3DSS, because we aim to define their taxonomy scheme.

In general, deep learning methods for 3DSS using point clouds could be separated into four classes, [8], [12], [13], [14]:

- Point Based Methods: Which use the raw point cloud directly as input for the 3DSS algorithms.
- Projection Based Methos: Which project the point cloud onto a 2D grid, then exploit mature 2D convolution techniques for 2DSS and finally project the labels back into 3D space.

- Discretization Based Methods: Which transform the point cloud to a discrete representation on which 3D convolution operator is applicable.
- Hybrid Methods: Which combine two or more techniques which belong to the previously described categories.

Hereafter we refer to the previously described taxonomy as the main taxonomy scheme just for comparison purposes. In the following paragraph, we present a connection between the different category names included in each review paper with the main taxonomy scheme.

Regarding the main taxonomy scheme, the (i) *Multi-View CNNs* and *Unordered Point Cloud Processing* [10], (ii) *Multi-View Based* [22], (iii) *Multiview Based* [18], (iv) Multi-View [7], (v), Image Based [9], (vi) *RGB-D Image Based* [23], (vii) *Dimensionality Reduction* [20] and (viii) *Multi view* [21] categories, are subcategories of Projection Based Methods or include some projection based methods along with others. Additionally, based on the main taxonomy scheme the (i) *Volumetric*, *Unordered Point Cloud Processing* and *Ordered Point Cloud Processing* [10], (ii) *Volumetric Methods* [22], (iii) *Voxel Based* [9], [18], [21], [23], (iv) *Voxel* and *Higher Dimensional Lattices* [7] and (v) *Methods Based on Voxelization* [20] categories, are subcategories of Discretization Based Methods or include discretization based methods along with others. Moreover, in respect to the main taxonomy scheme the (i) *Ordered Point Cloud Processing* [10], (ii) *Direct* [22], (iii) *Directly Process on Point Cloud Data* [18], (iv) *Deep Learning Directly on Raw Point Clouds* [7], (v) Point-Based Methods [9], [23], (vi) *Primitive Points*. [20] and (vii) Point Cloud Based [21] categories, are the same as Point Based Methods or include point based methods along with others.

Bello et al., 2020, Zhang et al., 2019 and R. Zhang et al., 2023 propose a more general taxonomy scheme for 3DSS deep learning methods, using only two broad categories i.e.,

- *Indirect* [22] , *Structured Grid Based Learning* [7] or *Rule-Based methods* [23] and,
- *Direct* [22], *Deep Learning Directly on Raw Point Clouds* [7] or *Point-Based methods* [23]

The authors used a different name to describe the same category of methods. To be more specific, the first group of methods transformed the given point cloud into a regular grid representation to apply the convolution operation on it, while the second group of methods use the point cloud directly as an input to the deep learning algorithms. Thus, the above scheme could be considered a superset of the main taxonomy scheme. Furthermore, R. Zhang et al., 2023 include *RGB-D Image Based* methods to their taxonomy scheme. This new subcategory is the core subcategory of Projection Based Methods in the main taxonomy scheme. We think that the name RGB-D Image Based methods it is easily confused with the RGB-D methods, presented in Griffiths and Boehm, 2019 review paper, which refer to the methods applied to data collected with RGB-D sensors e.g., in indoor environments. Hence Projection Based Methods seems to be a more relevant name for the category. Finally, A. Zhang et al., 2023 include the *Multiple Data Formats* category among others and divide it into *Multi Representational Methods* and *Multi-Modal Methods*. In the main taxonomy scheme the *Multi Representational Methods* is the same category as the Hybrid Methods while the *Multi-Modal Methods* for 3DSS could be classified following the same taxonomy scheme as the unimodal 3D Semantic Segmentation methods i.e., Point Based Methods, Discretization Based Methods etc. More concretely, multimodal 3DSS methods are classified with the same taxonomy scheme as unimodal 3DSS methods i.e., multimodal methods are not treated as a subcategory of unimodal 3DSS methods but as an independent category of methods.

**Proposed 3DSS taxonomy scheme:** After the examination of the previously described 3DSS review papers we found that, the Point Based methods of the main taxonomy scheme are usually included to categories like Direct Methods, Directly Process Point Cloud Data, Deep Learning Directly on raw Point Clouds, Methods based on primitive points etc. terms, but without any confusion about the methods included to these categories i.e., although the categories have a different name, they include almost the same methods. We chose the term Point Based Methods for our taxonomy scheme to classify these methods, because we think that describes better and compactly the general idea of the methods i.e., to extract meaningful for 3DSS features using as input the raw point cloud. Moreover, the Graph Based methods are commonly included to Point Based methods, in the literature. In our taxonomy Graph Based methods are a separate category, because feature extraction process is based on the nodes and the edges of graphs and not only based on the 3D points as in Point Based Methods.

Furthermore, we believe that Dimensionality Reduction Based Methods term introduced by A. Zhang et al., 2023 describes better the methods included in Projection Based Methods category in the main taxonomy scheme, because Projection Methods can also be used to describe a sub class of methods which use different projections e.g., Birds Eye View (BEV), cylindrical, polar etc. to perform 3DSS. Thus, we adopt Dimensionality Reduction Based Methods term to our taxonomy scheme. Additionally, Discretization Based Methods term describes better the methods which commonly referred as Voxel, Voxelization etc. as well as those commonly referred as Permutohedral Lattices or Lattices etc. because the main idea of these methods, is the creation of a new discrete representation based on the 3D point cloud and then the use of 3D convolution in order to extract meaningful features for 3DSS. Finally, Hybrid Methods category includes those methods which use a combination of techniques of the previously described categories.

To sum up, we use the following taxonomy scheme to classify 3DSS deep learning methods (Figure 1):

- Point Based Methods (Section 5.1. Point Based Methods):
  - Point Based Methods use the raw 3D points to extract meaningful features for 3DSS.
- Dimensionality Reduction Based Methods (Section 5.2. Dimensionality Reduction Based Methods):
  - Transform the 3D point cloud into a lower dimensional space e.g., images, perform semantic segmentation into that space and finally projects the labels back into 3D space.
- Discretization Based Methods (Section 5.3. Discretization Based Methods):
  - Transform the point cloud into a discrete representation without dimensionality reduction and then apply the convolution operation for 3DSS.
- Graph Based Methods (Section 5.4. Graph Based Methods):
  - Transform the point cloud into a graph and use the nodes and the edges of the graph to extract meaningful features for 3DSS.
- Hybrid Methods (Section Hybrid Methods):
  - Combine two or more from the previously described categories.

**Point Based Methods Taxonomy Scheme:** The Point Based Methods use the raw 3D points directly to extract meaningful features for 3DSS. In this subsection, the taxonomy scheme for the point-based methods is defined. Firstly, the Pointwise MLP category contains the point based 3DSS methods which apply a per point process using several shared Multi-Layer Perceptron (MLPs) to extract local features and then extracts the global information using an symmetrical aggregation function [7], [8], [9], [12], [14], [23]. In general, convolution is the fundamental operation of a broad category of 2D deep learning networks i.e., CNNs, with remarkable results. However, the implementation of convolution operation using 3D point clouds is not a straightforward process due to their characteristics. The methods which investigate the implementation of convolution operation directly onto the 3D point clouds are included into the Point Convolution category. Guo *et al.* (2021); Camuffo, Mari and Milani (2022) subdivide the convolution based methods which used directly onto 3D point clouds, into the continuous and discrete ones, based on the space that are applied. Gao *et al.* (2022); Zhang *et al.* (2023); Zhang *et al.* (2023); Jhaldiyal and Chaudhary (2023); Rauch and Braml (2023) include the 3D convolution based methods as a subcategory of Point Based methods but without a further categorization. Thirdly, Recurrent Neural Networks (RNN) treat the given data e.g., image, 3D point cloud etc., as sequences of features. Furthermore, RNN based methods aim to extract the contextual information between the data elements i.e., pixel, 3D points etc., by recalling the features gathered earlier in the sequence or by forgetting some of them. [8], [9], [12], [14], [21], [23]. The 3DSS methods which use RNNs are classified into the Recurrent Neural Networks category. Finally, the 3DSS methods which use the transformer architecture and attention mechanism, are included into the Attention Mechanism and Transformers category [8], [14], [20], [21], [23].

**Dimensionality Reduction Based Methods Taxonomy Scheme:** The Dimensionality Reduction methods, transform the 3D point cloud into lower dimensional space data e.g., images, apply semantic segmentation techniques using that data and finally projects and fuse the labels back into 3D space. In this subsection, the subcategories of the Dimensionality Reduction methods are defined based on the projection used for the dimensionality reduction of the 3D point clouds. To be more specific, the Dimensionality Reduction methods are further classified into the Multiview, Spherical and Birds-Eye View (BEV) and Multiple Projections categories. Multiview methods replicate the photography of objects, scenes etc. To be more specific, they use the 3D data as the scene and then capture many images artificially, around the data using a set of predefined viewpoints. Finally, the created images are used for 2DSS. Furthermore, the Spherical and BEV methods use the spherical and BEV projections respectively, to create the images which are finally fed into the 2DSS process [7], [8], [9], [10], [12], [13], [14], [18], [20], [21], [22]. Moreover, the Multiple Projection category includes the methods which combine different projections to perform the 3DSS. In general, the created images are called range images when they are produced using LiDAR data without the assigned colors from a camera. Finally, the Point Cloud Serialization methods, which are commonly used in combination with Discretization techniques, transform the given point cloud into a lower dimension regular structure which is furthered processed for 3DSS and so could be considered as a dimensionality reduction technique.

**Discretization Based Methods Taxonomy Scheme:** The Discretization Based Methods transform the 3D point cloud into a discrete representation without dimensionality reduction and then apply the convolution operation for 3DSS. The main idea of these methods is to handle the unorder property of 3D point clouds by transforming the 3D point cloud to an appropriate representation for convolution operation in 3D space. In this subsection the subcategories of the Discretization Based Methods are defined. Most 3DSS review papers [9], [10], [18], [19], [21], [21], [22] focus on voxel-based methods. However, other review papers classify the Discretization Based methods into the Dense and Sparse subcategories, while some of them also use a subcategory for the methods which transform the 3D point cloud into a higher dimensional lattice to finally perform the 3DSS [7], [8], [12], [14], [20]. However, the proposed taxonomy scheme includes the methods which transform the 3D point cloud into a higher dimensional lattice before 3DSS, into the Sparse subcategory. Dense Discretization methods ignore the distribution of points and discretize the entire space of the 3D point cloud using a 3D grid of specific size. Sparse Discretization methods take into account the distribution of 3D points and discretize only the occupied space resulting in more efficient algorithms in respect to execution time. In this effort, only the sparse methods are presented because they dominate the category nowadays.

**Graph Based Methods Taxonomy Scheme:** The Graph based methods, transform the point cloud into a graph and use the nodes and the edges of the graph to extract meaningful features for 3DSS. In this category there is no further categorization of the methods proposed. Graph based methods are commonly included as a subcategory of Point Based Methods. However, the proposed taxonomy scheme defines them as a separate class because they transform the point cloud into a new representation i.e., a graph, and are not applied directly on the initial 3D point cloud.

**Hybrid Methods Taxonomy Scheme:** When a developed algorithm uses two or more techniques of the previously described categories, it is classified into the hybrid methods category. In this subsection, a further categorization of the hybrid methods is proposed. In fact, the proposed taxonomy scheme for the hybrid methods takes into account the techniques which are combined to create the developed algorithm, resulting in eight subcategories the all methods, Discretization-Point-Reduction, Graph-Discretization, Graph-Reduction, Point-Discretization, Point-Graph, Reduction-Discretization, Reduction-Point. In fact, there are more than eight possible categories, but we define those eight based on the examined papers.

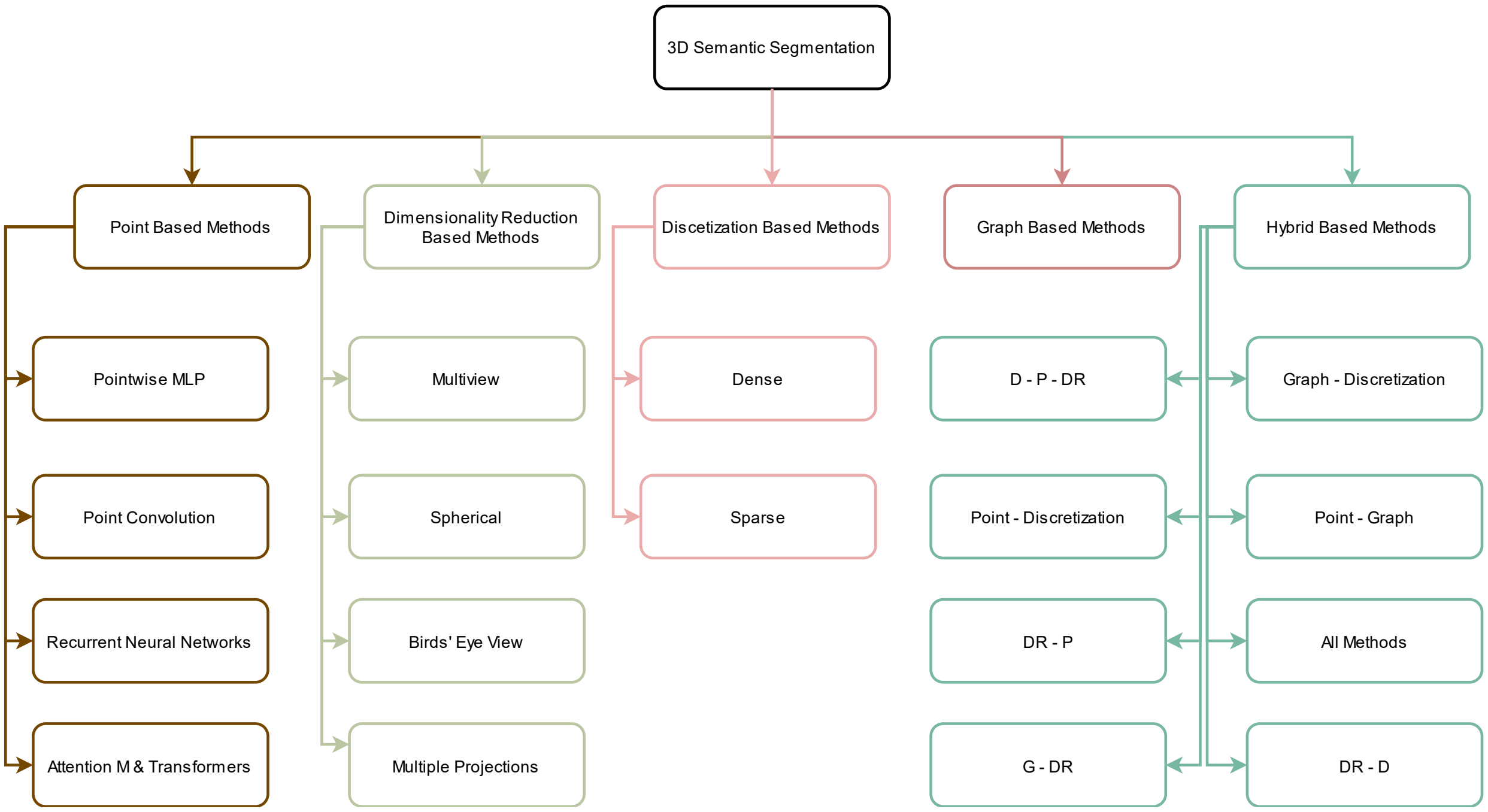

***Figure 1:*** *The proposed taxonomy scheme of the deep learning 3D Semantic Segmentation methods. Abbreviations: Attention Mechanism and Transformers (Attention M. & Transformers), Discretization Based Methods (D), Dimensionality Reduction Based Methods (DR), Graph Based Methods (G), Point Based Methods (P)*

## 3. 3D SEMANTIC SEGMENTATION DATASETS

In general, the benchmark datasets play a significant role to the proper investigation of the deep learning algorithms. The scope of this section is to provide a useful guide of the available 3DSS datasets along with their characteristics and their links. In 3DSS, the available datasets can be divided into indoor and outdoor ones. Thus, in Table 1 we summarize the available indoor datasets while in Table 2 we presented the available outdoor datasets for 3DSS.

***Table 1:*** *The indoor 3D Semantic Segmentation Datasets*

| **Dataset Name** | **Sensors** | **Sensor Type** | **Synthetic** | **#Classes** | **#Points** | **#Scans** | **Features & Notes** | **#Images** | **Mesh** | **Link** |
|---|---|---|---|---|---|---|---|---|---|---|
| 2D-3D-S S3DIS [24], [25] | Matterport Camera | RGB-D | NO | 13 | 695.878.620 | 5 | Surface normals, depth, 2D-3D instance - semantic annotations | 70.000 | Yes | http://buildingparser.stanford.edu/dataset.html |
| Freiburg Campus [26] | SICK LMS | Laser Range Scanner | NO | --- | 13.090.000 | 77 | --- | --- | --- | http://ais.informatik.uni-freiburg.de/projects/datasets/fr360/ |
| ViDRILO [35] | Microsoft Kinect | RGB-D | NO | --- | --- | 5 | images, 3D point cloud, semantic category, presence/absence of a list of 15 predefined objects | 22.454 | --- | https://www.rovit.ua.es/dataset/vidrilo/index.html |
| The Replica Dataset [34] | Custom System. 1 RGB, 1 IR, IMU, IR Pattern Projector, 2 Grayscale, | RGB-D | NO | 88 | --- | --- | geometry, normals, semantic and instance segmentation labels | --- | YES | https://github.com/facebookresearch/Replica-Dataset |
| SUN3D [33] | ASUS Xtion PRO LIVE sensor | RGB-D | NO | --- | --- | --- | 2D semantic labels, camera poses, Transfer annotations from 2D to 3d | --- | NO | https://sun3d.cs.princeton.edu/ |
| SUN RGB-D [32] | Intel RealSense, Asus Xtion and Kinect | RGB-D | NO | 800 | --- | --- | Transfer annotations from 2D to 3D | 10.335 | NO | https://rgbd.cs.princeton.edu/ |
| SceneNN [31] | Asus Xtion PRO - Microsoft Kinect | RGB-D | NO | --- | --- | --- | Transfer annotations from 2D to 3D | --- | YES | https://hkust-vgd.github.io/scenenn/ |
| SceneNet [30] | --- | RGB-D | YES | >250 | --- | --- | 3D scene labelling using voxels | --- | YES | https://robotvault.bitbucket.io/scenenet-rgbd.html |
| RGB-D Scenes Dataset v2 [29] | --- | RGB-D | NO | --- | --- | 14 | --- | --- | --- | https://rgbd-dataset.cs.washington.edu/dataset/rgbd-scenes-v2/ |
| ScanNet [28] | Structured Sensor on iPad Air 2 | RGB-D | NO | >50 | --- | 1513 | --- | --- | --- | http://www.scan-net.org/ |
| Matterport3D [27] | Matterport Camera | RGB-D | NO | 40 | --- | --- | 2D-3D annotations, surface reconstruction, camera poses | 194.400 | YES | https://niessner.github.io/Matterport/#paper |

***Table 2:****The outdoor 3D Semantic Segmentation Datasets. Abbreviations: Mobile Laser Range Finder (MLRF), Laser Scanner (LS), LiDAR (L), Photogrammetry (P), Mobile LiDAR (ML), Stereo Photogrammetry (SP), Aerial LiDAR (AL), Camera (C), Airborne Laser Scanner (ALS), Mobile Laser Scanner (MLS), Depth Camera (DC), Airborne Laser Radar (ALR), Multispectral LiDAR (ML),*

*Hyperspectral Camera (HC), Terrestrial Laser Scanner (TLS), Fisheye Camera (FC), Long-Range Radar (LRR), Mid-Range LiDAR (MRL), Short-Range LiDAR (SRL), Mobile Mapping System (MMS)*

| Dataset Name | Sensors | Sensor Type | Synthetic | #Classes | #Points | #Scans | Notes | Spatial Size | Mesh | Link |
|---|---|---|---|---|---|---|---|---|---|---|
| Freiburg, Pittsburgh, Wachtberg [36] | SICK LMS, Different Laser Scanners, Velodyne HDL64-E | MLRF, LS, L | NO | 13, 14, 5 | ≈13M | 77 | --- | --- | --- | http://ais.informatik.uni-freiburg.de/projects/datasets/fr360/ |
| Swiss3DCities[37] | Multicopter UAV | P | NO | 5 | ≈223M | --- | - 3 point clouds with different resolutions for each city (500k, 15M & 225M)<br>- nadir & oblique images<br>-RGB<br>-GSD 1.28cm | $\approx 2.7 \times 10^6 m^2$ | NO | https://zenodo.org/records/4390295<br>https://github.com/NomokoAG/Swiss3DCities |
| Paris-CARLA-3D [38] | Velodyne HDL32 | ML | YES | 23 | ≈700M (Synthetic)<br>≈60M (Real) | --- | -Both Real & Synthetic Data<br>-RGB (Ladybug5)<br>-Tilted LiDAR acquisition | ≈550m (R)<br>≈5.8km (S) | NO | https://npm3d.fr/paris-carla-3d |
| LiDAR-CS[39] | Velodyne Puck 16, 32E, HDL-64E, 128, ONCE-40, Livox | ML | YES | 5 | ≈1.8M | --- | -Labels only for Object Detection | --- | NO | https://github.com/LiDAR-Perception/LiDAR-CS |
| SMARS[40] | Synthetic Camera | SP | YES | 8 | --- | --- | -Multimodal -GSD 0.30 0.50 m<br>-RGB | 2Cities | YES | https://www2.isprs.org/commissions/comm1/wg8/benchmark_smars/ |
| SEMANTIC3D[41] | --- | LS | NO | 8 | ≈4000M | 30 | -RGB from cameras | --- | NO | http://www.semantic3d.net/<br>https://github.com/nsavinov/semantic3dnet |
| SensatUrban[42] | fixed wing UAV, SODA camera | P | NO | 13 | ≈3700M | --- | -3 cities (Birmingham, Cambridge, York (test)) | $\approx 7,6 km^2$ | NO | https://github.com/QingyongHu/SensatUrban |
| SWAN[43] | Ouster (64 channels) | ML | NO | 24 | ≈16000M (900M labeled) | --- | -Both Instance & Semantic Labels | ≈150km | NO | https://ieee-dataport.org/documents/swan-3d-point-cloud-dataset |
| RELLIS – 3D[44] | Ouster OS1, Velodyne Ultra Puck, 3D stereo camera, RGB camera | ML, SP | NO | 16 | ≈0.18M | --- | -2D & 3D annotations | --- | NO | https://github.com/unmannedlab/RELLIS-3D |
| DAPS3D[45] | Ouster OS0 (64-lines) | ML | YES | 4 | --- | --- | Both real & synthetic data | --- | YES | https://github.com/subake/DAPS3D |
| Hessigheim 3D (H3D) [46] | Riegl VUX-1LR, 2 RGB (Sony Alpha 6000) | AL, C | NO | 11 | ≈74.5M | --- | Multitemporal data, 800 pts/$m^2$ | --- | YES | https://ifpwww.ifp.uni-stuttgart.de/benchmark/hessigheim/default.aspx |
| Campus3D [47] | DJI Phantom 4 pro | P | NO | 24 | ≈0.95M | --- | 3D Labelling based on 2D labels | ≈1.58 $km^2$ | NO | https://github.com/shinke-li/Campus3D |
| Learnable Earth Parser [48] | --- | AL | NO | 2 - 4 | ≈100M | 7 | --- | ≈7.7$km^2$ | NO | https://romainloiseau.fr/learnable-earth-parser/ |
| OpenGF [49] | --- | ALS | NO | 2 | ≈550M | 9 | No RGB | ≈47$km^2$ | NO | https://github.com/Nathan-UW/OpenGF |
| Paris-Lille-3D [50] | Velodyne HDL-32E | ML | NO | 50 | ≈143M | --- | No RGB | ≈2km | NO | https://npm3d.fr/paris-lille-3d |
| DLA-Net [51] | 2 RIEGL VQ-450<br>4 high resolution cameras | MLS | NO | 8 | ≈158M | 6 | Facades | ≈3km | NO | https://github.com/suyanfei/DLA-Net |
| Toronto 3D [52] | Teledyne Optech Maverick (32 Lines) | ML | NO | 8 | ≈78M | --- | Urban Area | ≈1km | NO | https://github.com/WeikaiTan/Toronto-3D |
| Min3D [53] | Velodyne, Livox, Intel RealSense D455 | ML, DC | NO | --- | 46.138 point clouds | --- | Indoor & Outdoor | ≈0.63$km^2$ | NO | https://3dom.fbk.eu/benchmarks |
| MSNet [54] | SICK LMS291 LRF | MLRF, ALR | NO | 7 | ≈8.2M | --- | -Wuhan University<br>-5–10 points/m2 | --- | NO | https://github.com/wleigithub/WHU_pointcloud_dataset |
| N3C-California [55] | --- | AL, C | NO | 4 | --- | --- | Urban – Rural ≥8 pts/$m^2$<br>GSD 100 cm<br>LiDAR & Image | ≈725 $km^2$ | NO | https://github.com/wymqqq/IKDNet-pytorch?tab=readme-ov-file<br>https://pan.baidu.com/s/1iCe9YUlgcwweSCPnTBe0Vg#list/path=%2F |
| DFC 2018 [56] | Optech Titan MW (14SEN/CON340), ITRES CASI 1500 (S/N 2525), DiMAC ULTRALiGHT+ (Optech D-8900) | MSL, HC, C | NO | 20 | --- | --- | University of Houston<br>GSD 1m (HI)<br>GSD 5cm (VHR) | ≈4.3 km$^2$ | NO | https://hyperspectral.ee.uh.edu/?page_id=1075 |
| DublinCity [57] | Leica RCD30, NIKON D800E | ALS, C | NO | 4 - 6 | ≈1400M (260M Labeled) | --- | Urban ≈299 pts/m$^2$<br>GSD = 3.4cm | ≈2 km$^2$ | NO | https://v-sense.scss.tcd.ie/DublinCity/ |
| ArCH3D [58] | FARO Focus 3D X 130 & 120, Riegl VZ-400<br>Nikon D880E, D3100 & D3X, SONY Ilce 5100L, DJI Phantom 4 pro | TLS, UAV C, C | NO | 10 | ≈130M | --- | Cultural Heritage, Indoor & Outdoor | --- | NO | https://archdataset.polito.it/ |
| TerraMobilita/iQmulus [59] | Stereopolis II | MLS | NO | 4-9 | ≈300M | --- | Urban Environment | 10km | NO | http://data.ign.fr/benchmarks/UrbanAnalysis/index.html |
| KITTI-360 [60] | Velodyne HDL-64E, SICK LMS 200, | MLS, 2 FC, C | NO | 37 | ≈1000M | --- | Urban | 73.7km | NO | https://www.cvlibs.net/datasets/kitti-360/index.php |
| nuScenes [61] | Basler acA1600-60gc, Continental ARS 408-21, Velodyne HDL32E | 6 C, 5 LRR, L | NO | 23 | ≈1100M | --- | Urban | --- | NO | https://www.nuscenes.org/ |
| Oakland3D [62], [63] | Navlab11 SICK LMS | LS | NO | 44 | ≈1.6M | --- | University | --- | NO | http://www.cs.cmu.edu/~vmr/datasets/oakland_3d/cvpr09/doc/ |
| Paris-rue-Madame [64] | Velodyne HDL32 | ML | NO | 26 | ≈20M | --- | Urban | 0.16km | NO | https://people.cmm.minesparis.psl.eu/users/serna/rueMadameDataset.html |
| Waymo [65] | --- | 1 MRL, 4 SRL, 5 C | NO | 23 | ≈0.177M /frame | --- | Urban | 76 km$^2$ | NO | https://waymo.com/open/ |
| TUM-City-Campus [66] | Velodyne HDL64E | MODIS SA MMS, ML | NO | 8 | ≈1700M (40M annotated) | 8000 | Urban Campus | 0.2 km$^2$ | NO | https://www.pf.bgu.tum.de/en/pub/tst.html |
| SemanticKITTI [67] | Velodyne HDL64E | ML | NO | 25 (28) | ≈4500M | 23201/20351 | Urban | 73.7km | NO | http://www.semantic-kitti.org/ |

## 4. EVALUATION METRICS ON 3D SEMANTIC SEGMENTATION

In general, the performance of the developed algorithms is examined using different metrics regarding the under investigation task. Basically, every benchmark dataset provides a set of test data and evaluation metrics for comparison purposes. Every upcoming method which use the benchmark is evaluated using the data and the metrics. In fact, there are several metrics like accuracy, recall, precision and F1-Score. In 2D and 3D semantic segmentation the most common metric is the Intersection over Union (IoU) which better assess the performance of the model than the others. In this section, the commonly used evaluation metrics are described. But firstly, the confusion matrix should be examined.

**Confusion Matrix:** The confusion matrix is the basis on the evaluation of an algorithm using test data. It is created measuring the True Positives (TP), True Negatives (TN), False Positives (FP) and False Negatives (FN) for each class. In fact, such metrics are further used for the calculation of the evaluation metrics presented earlier. Commonly, the measured metrics i.e., TP, TN etc. are visualized using the Confusion Matrix. An example of a confusion matrix of two classes is presented below.

| Predicted Values | | Actual Values: Positive | Actual Values: Negative |
|---|---|---|---|
| | Positive | TP | FP |
| | Negative | FN | TN |

Comparing the predicted values with the actual values the TP, FP, FN and TN values could be derived forming the confusion matrix. While the above example is a 2 x 2 matrix, most of the time there are more than two classes e.g., n and hence the confusion matrix could be an n x n matrix, regarding the number of them.

**Evaluation metrics:** In general, the most straightforward metric is the accuracy i.e., the ratio between the correctly predicted values divided by the total number of values. However, there are different accuracy metrics presented in the literature i.e., the Overall Accuracy (OAcc), mean class Accuracy ($m_c$Acc) and mean Average Accuracy ($m_a$Acc). More concretely, the OAcc or Acc, is the predefined ratio for the test set (Eq 1), the $m_c$Acc is the mean accuracy of a class for a specific number of iterations (Eq 2) and the $m_a$Acc is the mean accuracy of all the classes divided by the number of classes (Eq 3). In fact, there is confusion between the different accuracy metrics, especially for multi class experiments. Hence, the reader should always examine the formulas used in any case.

$$\text{OAcc} = \frac{TP+TN}{TP+TN+FP+FN} \quad (1)$$

$$\text{m}_c\text{Acc} = \frac{1}{\#n}\sum_n OAcc \quad (2)$$

$$\text{m}_a\text{Acc} = \frac{1}{\#c}\sum_c OAcc \quad (3)$$

Where:

- TP, TN, FP, FN: True Positive, True Negative, False Positive and False Negative
- #n: Number of Iterations
- n: Iteration
- #c: Number of classes
- C: Class

In fact, the accuracy gives informative results when it is calculated using balanced data regarding the classes. Hence, there are more evaluation metrics derived using the confusion matrix, i.e., the precision, recall and F1-score. Intuitively, precision measures how vulnerable the model is in positive predictions or in other words the quality of the positive predictions, recall measures how prone the model is to correctly perform a positive prediction and F1-score is the harmonic mean between precision and recall. Another terminology about recall is true positive rate or sensitivity.

$$\text{Precision} = \frac{TP}{TP + FP} \tag{4}$$

$$\text{Recall} = \frac{TP}{TP + FN} \tag{5}$$

$$\text{F1-score} = 2\,\frac{Precision \times Recall}{Precision + Recall} \tag{6}$$

In most cases, the IoU metric is calculated to assess the performance of 2D and 3D semantic segmentation models. There are different IoU metrics like the mean IoU (mIoU) and the frequency weighted IoU. A different terminology of the IoU is the Jaccard Index. Intuitively, the IoU quantifies the degree of overlap between the predicted and the actual class e.g., of the masks. Moreover, the frequently weighted IoU takes into account the number of the data in each class e.g., in an urban area there will be less points representing the class bicycle than those represent the class street. Using the confusion matrix the IoU of each class can be calculated as follows.

$$\text{IoU} = \frac{TP}{TP + FP + FN} \tag{7}$$

$$\text{mIoU} = \frac{1}{\#c} \sum_{c} IoU \tag{8}$$

# 5. 3D SEMANTIC SEGMENTATION

## 5.1. Point Based Methods

**Pointwise MLP:** Up to the presentation of the PointNet [68] architecture, the extraction of 3D point cloud features was made manually. PointNet [68] was the first architecture which enabled the extraction of deep learning features directly using 3D point clouds. To be more specific, the input of the PointNet architecture was a point cloud with nx3 dimensions. Firstly, the authors introduced the T-Net, which learns to transform the given point cloud, e.g., rotate, translate and change its scale. Secondly, the transformed point cloud was fed into multiple shared Multi-Layer Perceptron (MLPs), resulting in a point cloud with nx64 i.e., local features dimensions which finally were transformed to nx1024 dimensions i.e., global features. Thirdly, the max-pooling operation was applied on the nx1024 point cloud selecting the maximum value between the 1024 values. The idea was that the maximum value was independent to the values order, thus it handles the unorder characteristic of point clouds. However, PointNet has several limitations, such as its inability to exploit the information of the local structures in point clouds. Hence, Qi et. al. (2017) proposed an extension of PointNet called PointNet++. The main contribution of PointNet++ was the proposed hierarchical structure which decodes the information of point clouds using three abstraction levels. The first one was implemented by the Sampling Layer, which finds the centroids of local regions by subsampling the given point cloud using the Farthest Point Sampling (FPS) algorithm. The second one was implemented by the Grouping Layer, which creates points neighborhoods using the Ball Query algorithm. Each point in the neighborhood was defined with relative coordinates to the neighborhood centroid, achieving the extraction of the neighborhood local characteristics. Finally, the PointNet++ architecture used several mini-PointNet layers to extract meaningful features from the point sets. PointNet++ used several abstraction levels for the feature extraction, i.e., in multiple scales, and thus achieved the extraction of both fine-grained and global features, resulting in an improved version of PointNet.

Since the introduction of PointNet and PointNet++, several methods have been proposed to improve the extraction of richer local information from point clouds, especially with computationally efficient approaches such as the RandLA-Net [70] architecture. Specifically, Hu et al. (2020) [70] tried to expand the application of pointwise MLP methods from small point clouds, e.g., thousands of points, to the large ones, e.g., millions of points, without using pre/post processing steps and by replacing the computationally heavy sampling methods with lighter ones. To be more specific, Hu et al. (2020) stated that the majority of methods presented so far used complicated sampling methods, categorized into (i) the Heuristic Sampling methods (Farthest Point Sampling (FPS), Inverse Density Importance Sampling (IDIS), and Random Sampling (RS)), and (ii) the Learning-based sampling methods (Generator-based Sampling (GS), Continuous Relaxation-based Sampling (CRS), and Policy Gradient-based Sampling (PGS)). Furthermore, the existing local feature learners were based on point cloud transformations, e.g., voxelization, graphs, and they were not decomposing the complicated local information of point clouds of highly detailed objects, resulting in time-consuming approaches with low performance using such data. The authors proposed the Local Feature Aggregation (LFA) module accompanied by Random Sampling to overcome the aforementioned limitations. More precisely, the LFA module included the Local Spatial Encoding (LocSE) and the Attentive Pooling (AP) submodules. The LocSE submodule firstly found the k-nearest neighbors for each center point using the Euclidean distance, then defined the relative point positions of the neighboring points to the center point, and finally concatenated the features of the neighboring points with the relative positions to create an augmented feature vector. The AP submodule was introduced on behalf of max/mean pooling approaches. It first received the created feature vector and computed attention scores using several shared MLPs and SoftMax. Finally, the important features were selected based on a weighted summation approach. In addition to the aforementioned submodules, the authors present the Dilated Residual Block (DRB), which consists of multiple LocSE, AP and Skip Connections in order to collect information using multiple receptive fields, preserving the local details of points' neighborhoods. In summary, the RandLA-Net architecture was constructed using several LFA modules and Random Sampling and evaluated on 3D semantic segmentation benchmarks, achieving state-of-the art results.

Fan et al. (2021) [1] proposed the SCF module aiming to provide a robust method for 3D semantic segmentation of large point clouds. The SCF module was composed by three main blocks: the Local Polar Representation (LPR), the Dual-Distance Attentive Pooling (DDAP), and the Global Contextual Feature (GCF). More precisely, LPR aimed to find a representation of the local neighborhood of points which was invariant to the Z-axis rotation. In fact, the objects belonging to the same category, e.g., chairs in a scene like an office, were presented with different orientations resulting to 3DSS features which were orientation-sensitive. Thus, using LPR, the neighborhood of each point was represented using polar instead of cartesian coordinates. More precisely, the points' initial local representation was defined using the Euclidean distance. Then, the local direction of each point in the neighborhood was extracted by calculating and then updating two angles $\phi$ and $\theta$. Specifically $\phi$ was defined among each 3D point in the neighborhood of the under-process 3D point, the 3D point, and the X-axis, in the XY-plane. Moreover, $\theta$ was defined among each 3D point in the neighborhood of the under-process 3D point, the 3D point and the Z-axis, in the XY-plane. $\theta$ and $\phi$ were then updated based on the barycenter point of each neighborhood. The LPR module answered the research question of *how to represent the local context of a 3D point cloud?* Afterwards, the DDAP module was proposed to learn local context features using the LPR representation of points neighborhood. To be more specific, the DDAP module used the geometric distance in the world space, the feature distance in the feature space, and geometric patterns. The feature distance was the mean value of the subtraction between the features of the central point and the features of the under-process neighboring point. Furthermore, the geometric patterns were created by concatenating the LPR representation with the original coordinates and feeding them into shared MLP layers. Afterwards, the MLP output was fed into SoftMax, forming the attentive pooling layer, which was learnable. Finally, the local contextual features were gathered using a weighted sum operation between the neighboring point features and the output of the attentive pooling layer. The DDAP module answered the research question of *how to learn local contextual features?* Apart from the local features, the 3DSS methods also need global features to achieve high-end results. To this end, the authors proposed the GCF module which combined the (i) local x, y, and z coordinates with (ii) a ratio, defined as the number of points into the local neighborhood divided by the number of points into the global neighborhood, using MLP. The GCF module answered the research question of *how to learn global contextual features?* Finally, the SCF architecture was defined using multiple SCF modules in an encoder-decoder structure.

Qiu, Anwar and Barnes (2021) [71] proposed a 3DSS network aiming to solve several limitations of the SotA approaches. To be more specific, the authors stated that the SotA methods were time-consuming, or they created intermediate representations like Graphs or Voxels, which cause a partial loss of information. To this end, they proposed a point-based method which aimed to remedy three major drawbacks: (i) ambiguity in close points, (ii) redundant features, and (iii) inadequate global representations. Firstly, ambiguity in close points refers to the limitations which occur due to the neighborhood selection process like overlapping regions or outlier selection. To alleviate the aforementioned disadvantage, the authors proposed an augmentation of the points' neighborhood. Secondly, redundant features refer to the models which use the same features multiple times to increase their performance. The authors stated that this redundant information increased the complexity of the models, rather than improving their performance. Hence, they proposed to categorize the model's input, e.g., geometric or semantic, in order to fully utilize them. Finally, inadequate global representations refer to the limitations of SotA features maps in regards of 3DSS. Specifically, the authors stated that SotA encoder-decoder methods decreased the details of the created feature maps, resulting in difficulties for 3DSS. Thus, the authors proposed a multi-resolution approach in order to preserve the fine-grained details through the model. The aforementioned improvements proposed by the authors were included into the Bilateral Context Module and the Adaptive Fusion Module. Firstly, the Bilateral Context Module was decomposed into Bilateral Augmentation, Augmentation Loss, and Mixed Local Aggregation. During the Bilateral Augmentation unit, the neighborhood of each centroid was defined using the kNN algorithm with 3D Euclidean Distance. Then the local context (geometric and semantic) of each centroid was defined by combining its absolute coordinates with the relative to the centroid coordinates for each neighboring point. The local context was augmented by using bilateral offsets, which were guided by the Augmentation Loss using penalties during the learning process. Finally, the Mixed Local Aggregation unit was used to construct a neighborhood representation which precisely demonstrated the local distinctness of neighborhoods. Secondly, the Adaptive Fusion Module was used to extract feature maps for 3DSS using multi-resolution features. Specifically, several point clouds with descending resolutions were processed using the Bilateral Context Module, extracting multiple scales of information. Then each feature map was gradually upsampled, to the original representation, forming an upsampled set of feature maps which were finally fused adaptively at point level. Overall, the authors proposed the BAAF-Net, which included the aforementioned modules, units and losses, and was evaluated using different benchmarks, metrics and training schemes.

**Point Convolution:** Li et al. (2018) [72] proposed the PointCNN architecture, aiming to handle the irregularity and unorder characteristics of point clouds for 3D learning applications. First and foremost, the authors provided a detailed analysis of the limitations regarding the extension of conventional 2D CNN to 3D space, which was the main research question of Point Convolution methods. Thus, they proposed to extend the application of the typical CNN on the 3DSS features rather than the 3D point cloud. PointCNN aimed to weight and permute the extracted features a predefined number of times, and then to use the convolution operation on them. To be more specific, PointCNN was fed with a set of 3D points, along with their corresponding features. X-Conv operator was proposed by the authors to be the main unit of PointCNN architecture. It was used to create a new representation, based on the input features, which had more depth in features and less points, e.g., it was a more abstracted and informative representation of the data than the first one. X-Conv for 3DSS exploited the furthest point sampling on an unorder local region of points which were first transformed from the global to a local coordinate system. Receptive field was crucial for 2D CNNs, contributing seriously to the performance of 2DSS algorithms. PointCNN defined the receptive field in 3DCNN by calculating a ratio (K/N) using the number of neighboring points of a 3D point in the current (K) and previous layer (N). To conclude, PointCNN was an encoder-decoder architecture which used the X-Conv operator in both the encoder and decoder part. The main difference was that in the encoder phase, the created representations had less points and richer features, while in the decoder phase they were the opposite, with an addition of skip connections.

Thomas et al. (2019) [73] presented the Kernel Point Convolution (KPConv) architecture which was inspired from the typical 2D convolution operator. Specifically, KPConv weights were carried by the 3D points in Euclidean space, in the same manner as the features. In general, the rigid KPConv operator was applied on the 3D points close to the convolution location. To be more specific, the neighborhood of the convolution location was defined using a specific radius. KPConv operator took as an input the neighborhood of points with coordinates relative to the convolution location. The convolution kernel was defined by points which were linked with their weights and aimed to transform the features of the given neighborhood points to a new set of features. Specifically, the transformation of features dimensions was done using a correlation function between the input and the kernel points. The authors used the Linear correlation function. Additionally, they presented a comparison between the Gaussian correlation and the Linear correlation to support their choice. Moreover, the authors proposed a deformable version of the KPConv operator, which was to consider the local geometry of the points by applying different shifts at the convolution location. In fact, the convolution location was critical for the proposed operator. Furthermore, the authors included grid subsampling to handle the varying density issue of some 3D point clouds, explained the Pooling Layer, the KPConv layer, and then analyzed the selection of the network parameters. Finally, the authors proposed two architectures. The first one was about classification, while the second one was about segmentation. To be more specific, the segmentation architecture included the classification architecture into the encoder part, while the decoder part applied nearest upsampling in combination with skip connections between the encoder and decoder features. To conclude, the proposed KPConv architectures were evaluated in different benchmarks for classification and segmentation purposes.

Liu et al. (2021) [74] introduced the FG-Net, a general point-based 3D deep learning architecture which can be used in different downstream applications like 3DSS or 3D classification. First and foremost, the authors presented several limitations of the SotA methods. For example, they contained time-consuming operations like Farthest Point Sampling, or intermediate representations such as graphs or voxels, they could not be applied to large-scale point clouds and had difficulty understanding detailed geometry. To address the limitations of existing methods, the authors proposed the FG-Net architecture in which the core module was called FG-Conv. Specifically, the FG-Conv module was composed of three main parts: Pointwise correlated features mining (PFM), Geometric Convolutional Modelling (GCM), and the Attentional Aggregation (AG). Before the PFM, the authors proposed a Noise and Outliers Filtering (NOF) approach. To be more specific, the neighborhood of each point was determined based on a radius NN ball query algorithm. Afterwards, an estimation of the point's neighborhood density was calculated. If the estimated density was lower than a threshold, the point was characterized as isolated and was deleted. Otherwise, the neighborhood of points was modeled as a Gaussian Distribution (μ, σ). Finally, neighborhood points were removed if the mean distance was not into the confidence interval of the Gaussian Distribution. After, NOF the point cloud was represented using the 3D coordinates along with the associated features like normals, colors, and learnt latent features. Using the filtered point cloud and knowing the neighborhood of each point, the PFM defined a similarity score, in both 3D and features space, between the center point and each neighboring point by calculating the inner product between them. Afterwards, the calculated similarity vector was passed into a learnt attention mechanism to transform the similarity scores based on the applied downstream 3D task, e.g., 3DSS. The new similarity scores were multiplied elementwise, with the 3D points resulting to the augmented attentional feature matrix which was concatenated with the original features. PFM aimed to construct a feature space in which the relevant features were enhanced while the irrelevant features were attenuated. Moreover, the GCM mimics the deformable convolution from 2D space to 3D point cloud data and aimed to model the geometric structure of points. Firstly, a correlation function was defined to calculate the association between the convolution kernel points and the local geometry, i.e., the points neighborhood in a local coordinate system. The correlation function aimed to overcome the unstructured and unorder characteristics of 3D point clouds by increasing their value as the kernel points were closer to the local geometry. Then, the kernel function was defined as the sum of the correlation values with learnable weights and applied to capture the local geometry. Finally, the Attentional Aggregation component was defined to decrease the information loss when using the geometric and features patterns by finding the meaningful features and aggregating them. The FG-Net was an encoder-decoder based architecture which leveraged multi-resolution point clouds. To be more specific, the encoder was based on residual learning blocks (RLB) inspired by ResNet [75]. In fact, the FG-Conv operator was the core of the RLB. Between the encoder and the decoder there was the Point Clouds Global Feature Extraction, which aimed to capture the global dependencies in point clouds using the Nonlocal Attentive Module. Finally, they proposed a learning sampling approach called IGSAM, using the advantages of inverse density (IDS) and Gumbel SoftMax sampling instead of the time-consuming point sampling methods. To conclude, the FG-Net was evaluated over different benchmarks for 3D classification and 3DSS using the mIoU metric. Additionally, the authors evaluated the performance of different models on 3DSS regarding their sampling technique, as well as the computation and memory consumption.

**Recurrent Neural Networks:** Huang, Wang and Neumann (2018) [76] observed that the local information incorporated into the 3D point clouds was not sufficiently exploited from the SoTA methods. Additionally, they stated that the SoTA methods spent a wealth of time on the computation of the local dependencies. Thus, they proposed the RSNets, a series of a Slice Pooling Layer (SPL), Recurrent Neural Network (RNN) Layers, and Slice Unpooling Layer (SUL). The aforementioned components were incorporated into the Local Dependency Module. Firstly, the under-process 3D point cloud was fed into the Input Feature Block (IFB). More concretely, the IFB transformed the unordered 3D point cloud into a set of unordered features. The unordered features were then input into the SPLs. More specifically, there were three SPL layers, one for each direction, i.e., x, y and z, called slices. Each slice contained a set of 3D points. Afterwards, a global feature vector for each slice was produced by aggregating the slice points' features. Finally, a set of an ordered sequence of features was created, which were the input of the RNN layer. In fact, the SPL created a representation of the features which can be exploited by the RNN bidirectional layers. The output features of the RNN layers were fed into the SUL to assign them to each point. To conclude, the RSNets architecture was evaluated using different large-scale point cloud benchmarks after a thorough analysis of ablation studies and experiments.

**Attention mechanism and Transformers:** In general, the attention mechanism included in transformer architecture is promising for 3DSS, as it is independent of the 3D point cloud characteristics such as irregularity, disorder [14], [23]. After the SotA performance of transformers on different tasks [77], Zhao et al. (2021) [78] stated that the self-attention mechanism appears particularly relevant to be used in 3D tasks, and proposed the Point Transformer architecture. First and foremost, the authors stated that the self-attention mechanism can be categorized into scalar and vector attention. In fact, the difference between the scalar and vector attention is the creation of scalar and vector scores, respectively. Thus, the vector attention mechanism was selected for the Point Transformer architecture due to its ability to capture more detailed information about neighboring 3D points [78]. Vector attention equation components were the three pointwise feature transformations ($\phi$, $\psi$, $\alpha$). Like MLPs or Linear Projections, the position encoding function $\delta$, a normalization function $\rho$ like SoftMax, the relation function $\beta$ and the mapping function $\gamma$. To be more specific, the vector attention mechanism on the proposed Point Transformer Layer (PTL) was applied to a local neighborhood of points defined using the kNN algorithm on a specific location, using the subtraction as the relation function, an MLP, two linear layers, and ReLU nonlinearity as the mapping function and position encoding $\delta$ into both functions $\gamma$ and $\alpha$. Furthermore, positional encoding was a very important component of transformer architecture, since it was the counterpart of convolution and recurrence operations [77]. Specifically, in Point Transformer, the positional encoding function was a trainable function based on the 3D point coordinates. Basically, positional encoding was an MLP with two linear layers and ReLU nonlinearity aimed to define the neighboring points' intra relationships. The aforementioned units consist of the Point Transformer Layer, which was the core of the Point Transformer Block. In fact, the Point Transformer architecture was created in encoder-decoder fashion, with skip connections between the encoder layer and their corresponding decoder layers. More concretely, the encoder was constructed using the Point Transformer Block and the Transition Down Layer, while the decoder the Point Transformer Block, and the Transition Up Layer. The Transition Down and Transition Up aimed to find a subset and superset of the input points, respectively. In 3DSS the Output Head mapped the decoder output to the predicted class using an MLP. To conclude, Point Transformer architecture was evaluated using different experiments in multiple tasks like 3DSS, Classification, and different metrics.

Wu et al. (2022) [79] stated that the Point Transformer architecture increases the number of channels and the weight encoding parameters as it goes deeper, resulting in overfitting and restricting the model to go deeper. To overcome the aforementioned problem, the authors introduced a new attention mechanism called Group Attention (GA), instead of the vector attention which was used in the Point transformer architecture. The GA was characterized as a more general case of the vector aimed to reduce overfitting and enhance the generalization of the model. The authors stated that neighborhood attention performs better than shifted-grid attention. To be more specific, gathering the local neighborhood of points using the kNN function outperforms the methods which construct the local neighborhood using uniform non-overlapping cells due to the different point density of 3D point clouds. Additionally, the authors tried to better reveal the 3D points relationships by adding an additional positional encoding mechanism aiming to fully exploit the geometric knowledge encapsulated into the 3D point coordinates. In general, the traditional based pooling procedures did not consider point density and overlapping. Hence, they proposed an improved pooling operation, using uniform grid partition to replace the commonly used pooling approaches such as FPS or Ball Query. Furthermore, the authors presented extended experiments and ablation studies to strengthen the advantages of the proposed architecture. To conclude, the Point Transformer V2 was evaluated using different benchmarks and metrics under different 3D tasks like 3DSS and 3D Shape Classification.

***Table 3:*** *Mean Intersection over Union (mIoU) and Overall Accuracy (OA) for Different Benchmark Datasets for the Point Based Methods*

| Algorithm | Year | Stanford3D | | Semantic3D | | ScanNet | | SemanticKITTI | | S3DIS | | SenSat Urban | |
|---|---|---|---|---|---|---|---|---|---|---|---|---|---|
| | | mIoU | OA | mIoU | OA | mIoU | OA | mIoU | OA | mIoU | OA | mIoU | OA |
| PointNet | 2017 | 47.71 | 78.62 | 14.6 | --- | 55.7 | 52.6 | 14.6 | --- | 47.6 | 78.6 | 23.71 | 80.78 |
| PointNet++ | 2017 | --- | --- | 63.1 | 85.7 | | 84.5 | 20.1 | --- | 54.5 | 81.0 | 39.97 | 84.30 |
| RandLA-Net | 2020 | --- | --- | 77.4 | 94.8 | 64.5 | --- | 53.9 | 88.8 | 70.0 | 88.0 | 52.69 | 89.78 |
| SCF-Net | 2021 | --- | --- | --- | --- | --- | --- | --- | --- | 71.6 | 88.4 | --- | --- |
| BAAF-Net | 2021 | --- | --- | 75.4 | 94.9 | --- | --- | 59.9 | --- | 72.2 | 88.9 | --- | --- |
| PointCNN | 2018 | --- | --- | --- | --- | 45.8 | --- | --- | --- | 65.4 | 88.1 | --- | --- |
| KPConv | 2019 | --- | --- | 74.6 | 92.9 | 68.6 | --- | 58.8 | 90.3 | 70.6 | --- | 57.58 | 93.20 |
| FG-Net | 2021 | --- | --- | --- | --- | --- | --- | --- | --- | 70.8 | --- | --- | --- |
| RSNets | 2018 | --- | --- | --- | --- | --- | 76.5 | --- | --- | 56.5 | 85.7 | --- | --- |
| PTv1 | 2021 | --- | --- | --- | --- | --- | | --- | --- | 73.5 | 90.2 | --- | --- |
| PTv2 | 2022 | --- | --- | --- | --- | --- | 75.2 | --- | --- | 71.6 | 91.1 | --- | --- |

## 5.2. Dimensionality Reduction Based Methods

**Multiview:** Tatarchenko et al. (2018) [80] proposed the tangent convolution in order to perform 3DSS using point clouds with millions of points. In fact, the tangent convolution approach was an extreme case of the multiview CNNs proposed by Su et al. (2015). The tangent convolution was proposed for dense prediction tasks instead of the shape recognition task which multiview CNNs remedy. Tangent convolution can be used upon different types of 3D data, e.g., mesh, point clouds, polygon soup, with the only constraint being the ability to estimate the normal vectors using them. In general, the authors stated that the 3D data, captured using different sensors, represent 2D structures embedded in 3D space and thus tangent convolution was based on the concept that the data was drawn from local Euclidean surfaces. To be more specific, tangent convolution first defined a tangent plane around every point and projected the local geometry on it, creating a tangent image, e.g., the tangent plane was exploited as an orthogonal to the point's neighborhood, virtual camera i.e., along its normal vector. Moreover, for each 3D point the orientation of the tangent plane was derived by covariance analysis of its local neighborhood. The local neighborhood was defined using a radius around the under-processing point. Afterwards, the covariance matrix of the neighborhood was estimated, resulting in the estimation of the neighborhood normal vector, i.e., the eigenvector of the smallest eigenvalue, and the x, y axes of the tangent plane i.e., the other two eigenvectors. However, to create the virtual images using the tangent planes, the point signals must be used to estimate image signals. The neighborhood points were projected onto the tangent plane of the under-process 3D points, resulting in a set of projected points. Hence, the projected points were a sampling of the continuous image space and so the authors investigated different interpolation approaches, e.g., Nearest Neighbor, full Gaussian mixture, or Gaussian mixture with top 3 neighbors to formulate the final tangent images. They stated that the sophisticated interpolation methods did not result in a significant improvement on the tangent images, and thus they proposed to use simple nearest-neighbor interpolation. In practice, the continuous space was replaced by a discrete space i.e., grid, resulting in the tangent images used to perform semantic segmentation. Furthermore, the authors proposed a UNet-like network for the semantic segmentation of the tangent images. First, they defined the core network operations, e.g., pooling and unpooling. To be more specific, the pooling operation was defined via hashing the 3D points onto a progressively coarser 3D grid with a predefined grid resolution, using modular arithmetic on individual point coordinates. More precisely, the points which were hashed together into the 3D grid, were used to pool their signal. Hashing deals with the irregularity property of 3D point clouds. Unpooling was performed reusing the hash indices of the pooling operation. Additionally, they proposed the Local Distance Features as the mean distance between the neighborhood points and the tangent plane. The Local Distance Features were used to create distance images which were exploited as an additional channel of the data. The proposed encoder-decoder architecture had two pooling layers and two unpooling layers respectively, using 3x3 kernels followed by Leaky ReLU with skip connections, while the last layer exploited 1x1 convolutions to assign the final classes. The pixel size of the tangent images along with the radius used to define the points neighborhood were used to define the receptive field of the convolutional layers. To conclude, the authors evaluated their algorithm using different benchmark datasets and metrics.

**Spherical:** Wu et al. (2017) [82] introduced the SqueezeSeg architecture, which utilized the 3D-2D spherical projection in combination with mature 2D semantic segmentation techniques to achieve real-time high-end results in 3DSS of road objects. First and foremost, the authors described the core steps of the existing 3DSS SotA approaches as the Ground Filtering, Points grouping, Hand-crafted feature extraction for each group, and Group classification steps. However, they mentioned that the existing ground removal approaches were characterized by poor generalization, time-consuming post-processing steps, or they relied on iterative algorithms like RANSAC [83], which depended on the quality of the random initialization. Additionally, the multi-stage existing methods included limitations, e.g., error aggregation phenomena. Thus, the authors proposed a dimensionality reduction learning algorithm which was based on a combination of 2D convolutional neural networks (CNNs) and conditional random fields (CRF). The general idea was to transform the input LiDAR point cloud into a compact representation, feasible for the convolution operation to extract the semantic labels and then to be refined using a CRF. In fact, the 3D point clouds properties made it difficult to apply convolution directly on them. Hence, the authors proposed to project the input 3D point cloud onto a sphere to create a dense 2D grid-base, similar to the ordinary image's representation. The created spherical projection had five features for each point, i.e., 3D cartesian coordinates, intensity and range. Moreover, the proposed network structure was inspired by the SqueezeNet [84] 2D architecture, which investigated the reduction of the AlexNet [85] parameters while preserving the performance of it. The SqueezeSeg encoder-decoder architecture was based on the fireModules and fireDeconvs layers instead of the traditional convolution, deconvolution layers, aiming to reduce the parameters of the model. Specifically, the fireModule input was the spherical projected 3D point cloud. Then, the spherical image was fed into a 1x1 convolution, resulting in a reduced size of feature channels. Afterwards, a parallel application of 3x3 and 1x1 convolutions were applied to recover the channel's size. The first layer was called the squeeze layer, while the second one was called the expand layer. The fireDeconvs, were the same as the fireModules, but with a deconvolution layer in between the squeeze layer and the expand layer. In general, 2DSS label maps suffer from blurry regions, especially between different classes, due to the down-sampling operation. Thus, the authors introduced a CRF to refine the produced label maps. More concretely, they proposed an energy function which has a unary potential term and a binary potential term, along with other terms. The latter was introduced as a punishment for labeling similar points with different labels. More precisely, the binary potential term was defined using two Gaussian Kernels, the former used both the spherical and cartesian coordinates, and the latter used only the spherical coordinates. Additionally, the Gaussian Kernels contained a set of empirical parameters. Overall, the labels' refinement process was made by trying to minimize the aforementioned energy function, using the mean-field iteration algorithm defined as an RNN. The refined labels were finally transferred into 3D space resulting in the real-rime 3DSS of road objects. To conclude, the SqueezeSeg architecture was trained using both real and synthetic LiDAR data and evaluated using different metrics.

However, Wu et al. (2018) [86] stated that the SqueezeSeg algorithm needed an improvement regarding the accuracy in order to be applied in real-world scenarios, while the manual creation of 3D training data for 3DSS was an extremely tedious process. To this end, Wu et al. (2018) [86] proposed the SqueezeSegV2 algorithm to firstly improve the performance of SqueezeSeg, and secondly to investigate the improvement of the synthetic training data creation. To be more specific, the authors stated that the accuracy degradation was mainly due to the dropout noise of real LiDAR data caused due to several circumstances like limited sensing range or mirror reflection. To alleviate the accuracy degradation, the authors proposed the Context Aggregation Module (CAM), a CNN module which leverages larger receptive fields in order to aggregate contextual information, and thus to be robust against missing points. Additionally, the authors changed the cross-entropy loss with the focal loss to handle the imbalanced distribution of point categories throughout the LiDAR point cloud. Furthermore, the authors enriched the created spherical images with an extra channel indicating if a pixel is missing or existing, which improved the segmentation accuracy. Additionally, batch normalization was included in the SqueezeSegV2 architecture to handle the internal covariate-shift phenomenon. In addition to the accuracy improvement, the authors tried to deal with the domain-shift problem. To be more specific, domain shift referred to the phenomenon in which the NN was trained into a different domain than to the applied domain, i.e., the source domain was different to the target domain, and so the generalization was poor. To alleviate the domain-shift problem, SqueezeSegV2 synthetic data render an intensity channel in addition to the rest by training a neural network in a self-supervised manner, to take point coordinates and depth as an input, and output the intensity values, aiming to mimic the real-world data. Additionally, they proposed geodesic correlation alignment and progressive domain calibration to reduce the gap between the source and the target domain. Specifically, on each training step, the SqueezeSegV2 network was fed with both synthetic and real data. Furthermore, they compute the focal loss on the synthetic batch to capture the semantic information and the geodesic distance of the output distributions of the synthetic and the real batch, to reduce discrepancies between the statistics of the source and the target domain. In progressive domain calibration, each layer of the network was calibrated progressively, from the first to the last, without the previous layer impacting the others. To be more specific, for each layer output statistics were calculated. Afterwards, the output mean was re-normalized to 0, and the standard deviation to 1, and in parallel the batch normalization parameters were updated with the new statistics. Overall, the new components were evaluated by the authors with ablation studies resulting in better performance than the SqueezeSeg architecture.

Milioto et al. (2019) [87] stated that the 3DSS LiDAR only SotA methods were time-consuming, while their models did not have enough representational capacity i.e., the number of neurons and learnable parameters was low. In addition to the time-consuming SotA approaches, the authors stated that there were not enough publicly available datasets for 3DSS, and that the SotA methods, e.g., [68], [69], [80], [88] cannot be applied in real-time scenarios using large scale point clouds. Thus, the authors proposed the RangeNet++ architecture, which utilizes spherical projection to facilitate fast scene assessment and decision-making of autonomous machines. In fact, the authors claimed that due to autonomous vehicle movement, the produced point clouds were affected by skewing i.e., the same as rolling shutter effect in images. Thus, they de-skewed the LiDAR point cloud before the projection. The spherical projection output was an enriched range image representation, e.g., tensor, which contained the x, y, z, range and remission information. Afterwards, the authors proposed an hourglass 2DSS architecture i.e., an encoder – decoder to perform 2DSS using the created representation. To be more specific, the encoder was a modification of the Darknet [89] backbone, e.g., to accept five channel images instead of three. In general, 2DSS experience difficulties, e.g., blurry areas, especially in object's borders. Additionally, the projection of the 3D point cloud onto a 2D plane resulted in a loss of information. Other methods, e.g., SqueezeSeg, used CRF to improve the 2DSS output and hence to improve the 3DSS, as described earlier. However, the authors of RangeNet++ stated that the improvement of the 2DSS using a CRF did not automatically result in a better 3DSS, especially due to the method applied to recover the 3D labels using the 2D ones, formulated as the label re-projection problem. To be more specific, each pixel of the 2D representation had a corresponding label, however multiple 3D points were assigned to each pixel of the image, resulting in assigning the same label, which was not accurate. Additionally, the label re-projection problem effect increased when using images with smaller resolution, which were crucial for real-time applications. To recover the entire information using the created 2D representation, the authors indexed them with the corresponding image coordinates to every point of the 3D point cloud resulting in a loss-less recovery of the 3D labels. Furthermore, the authors proposed a GPU-based kNN algorithm to cope with the label re-projection error. More concretely, for each point used to create the 2D spherical image, a window representing its neighborhood was empirically defined. Afterwards, the indices stored during the creation of the 2D representation were exploited to extend the neighborhood to contain the entire set of the range neighborhoods. Then, the range readings of the central line were replaced from the unwrapped to the real ones. This representation was the key of RangeNet++ to achieve real-time performance. An analogous matrix representation was constructed for the labels. Afterwards, they created a matrix which represented the range difference between the query point and its neighborhood points. Finally, the k nearest points which voted for the label of the query point were collected by using the inverse Gaussian kernel, argmin operation and cut-off thresholding. The aforementioned post-processing kNN-based approach was adopted from many upcoming architectures. To conclude, the authors evaluated the RangeNet++ algorithm using metrics such as border-IoU, as well as in-detail ablation studies and figures.

Xu et al. (2020) [90] described the issue of spatially-varying feature distribution of LiDAR images in detail. More concretely, the authors stated that the features distribution along the LiDAR images were different from the features distribution of the RGB images, especially for the SemanticKITTI [67] dataset due to the spherical projection applied on the data, resulting in poor performance of convolution operator. To be more specific, the feature distribution of LiDAR images was not identical across the image, while some features may exist only in local image regions. Additionally, the authors visualized the mean activation value of different layers of the RangeNet++ architecture depicting the sparse filter activation across the image. To this end, Xu et al. (2020) presented the SqueezeSegV3 architecture which utilized the Spatially-Adaptive Convolution (SAC) operator. On the one hand, the traditional convolution operator did not change the kernel weights across the image. On the other hand, SAC behaves differently on each part of the image, as the filters follow the features variations. The authors proposed different variations of SAC. SqueezeSegV3 architecture was based on the RangeNet++ [87] architecture, using a multi-layer cross entropy loss aiming to use features with semantic meaning. To conclude, using different variations of SAC, the authors achieved better performance on the 3DSS of many classes using the SemanticKITTI dataset.

Cortinhal, Tzelepis and Aksoy (2020) [91] stated that the SotA 3DSS methods predicted 3D labels without calculating uncertainty measurements, and thus, they proposed the SalsaNext architecture for real-time uncertainty-aware 3DSS of LiDAR point clouds. In line with the RangeNet++ [87] architecture, SalsaNext exploited the same spherical representation, but using intensity instead of remission. Furthermore, they introduced an improved SalsaNet [92] architecture to perform 2DSS using the proposed spherical representation of the 3D point cloud. More concretely, they proposed a Context Module in the first layers of the encoder, a Residual Dilated Convolution Block instead of traditional ResNet blocks, a Pixel-Shuffle Layer, a Central Encoder-Decoder Dropout operation, and an Average Pooling Downsampling layer instead of strided convolution to allocate less memory than SalsaNet. The Context Module had multiple dilated convolutions i.e., with different kernel size, which fused various perceptive fields in different scales, aggregating global context information. Furthermore, the pixel-shuffle layers were introduced instead of transposed convolution layers during upsampling to decrease the computation complexity of the network. Finally, they added dropout after both the encoder and decoder layers, except for the first and the last layers, resulting in improved networks performance. Furthermore, the authors provided an in-depth analysis of uncertainties, divided them into the aleatoric, which refers to the data uncertainty, and the epistemic, which refers to the model uncertainty. Moreover, aleatoric uncertainty can be divided into the homoscedastic, which refers to the aleatoric uncertainty which was independent to the different types of input data, and the heteroscedastic, which refers to the uncertainty that depends on them. If the sensor noise characteristics were known, a modified NN using assumed density filtering can be used to provide predictions along with their aleatoric heteroscedastic uncertainties. Finally, the authors estimated the epistemic uncertainty using Monte Carlo sampling during inference. In fact, the main drawback of dimensionality reduction methods was the information loss due to projection from 3D to a lower space. The authors used the kNN-based post-processing technique proposed in RangeNet++, applied during inference to cope with the aforementioned drawback. To conclude, the authors provided informative visualizations of the epistemic and aleatoric uncertainties on the images of the SemanticKITTI dataset in addition to the detailed ablation studies and evaluation of the SalsaNext architecture.

Xiao et al. (2021) [93] observed that the dimensionality reduction SotA methods were usually creating multi-channel images by stacking position channels, depth, intensity, and remission, commonly creating five-channel images, and processed them simultaneously without considering the distinct characteristics of each modality. Thus, the authors proposed the FPS-Net, aiming to handle the aforementioned modality gap problem by decomposing the five-channel images into three modalities and thus improving the performance of 3DSS applications. Moreover, they mentioned that the naïve stacking of pixel values along channels with different distribution may result in the models being trained on modality-agnostic features. To this end, they proposed a Modality Fused Convolution Network, which firstly learned modality-specific features and finally fused them using a high dimensional feature space representation in an encoder-decoder fashion. More specifically, modality-specific features were encoded by using multiple receptive field residual dense block (MRF-B) and decoded by using the recurrent convolution block (RCB). In detail, the MRF-B was composed of multidimensional convolutions, e.g., 1x1, 3x3, along with concatenation, batch normalization and ReLU, while the RCB was a recurrent neural network with 3x3 convolution, addition operation, batch normalization and ReLU. The modal-specific high dimensional features produced by MRF-B were concatenated and fed into a 1x1 convolution operation. After several MRF-B and downsampling layers, the features were passed into multiple upsampling and RCB layers. Moreover, the classifier output the predictions. Finally, a post-processing approach similar to RangeNet++ [87] was adopted to provide the final 3D labels. To conclude, the authors provided an evaluation of the proposed algorithms along with ablation study and experiments on well-known benchmark datasets.

Li et al. (2021) [94] mentioned that the SotA methods were characterized by very expensive operations, especially for applications requiring embedded platforms. They had low accuracy, or they included millions of parameters, hence they proposed the Multi-scale Interaction Network (MINet) to handle them. They introduced a lightweight network that included multiple scale paths, each of which extracted features of different levels, e.g., low or high, regarding the scale of the input image. Additionally, the top scale paths were densely connected with all the lower scale paths. However, the authors proposed a computation strategy to avoid redundant computations. In line with the RangeNet++ [87] architecture, the authors created 5D spherical projection images, i.e., x, y, z, depth and remission as the input to the MINet architecture. Furthermore, MINet architecture was composed by three modules: the Mini Fusion Module (MFM), Multi-scale Interaction Module (MIM), and the Up Fusion Module (UFM). Apart from the three modules, MINet architecture included two different blocks, the MobileBlock and the BasicBlock. The former has fewer parameters, as it utilized depth-wise convolutions, while the latter was more expensive. However, the MobileBlock was usually exploited using high-resolution images, aiming to extract detailed features, while the BasicBlock exploited using lower-resolution images, aiming to extract more abstracted features, to equalize the computation complexity. Firstly, the input 5D image was imported into the MFM. However, due to the different feature distribution of each modality, including x, y and z coordinates as separate modality, they were mapped into a different feature space using the convolution operation. Then, the features created from each modality were concatenated and fused using many MobileBlocks. After MFM, the MIM was applied. Specifically, MIM included three paths, the top, middle and bottom, with different scales. In each path the scale was decreased, applying the average pooling operation while the receptive field was increased, to gather more abstracted features. The authors showed that the previously described strategy, i.e., the decreasing resolution along with the increasing receptive field among the scale paths, resulted in efficient operations. Furthermore, the extracted feature maps on each scale firstly were resized using average pooling and then were passed to all the lower paths, allowing subsequent scale paths to focus on features that had not been extracted yet. Lastly, UFM fused the features extracted from the first layer of MFM to gather low-level spatial information, and each path of MIM to gather multi-scale information. Then the fused features were upsampled to the original resolution, further processed and added. Finally, the 2D predictions were re-projected back into 3D space. To conclude, the authors included several experiments and comparisons among the SotA methods and their own methods by retraining each SotA method from scratch and by using different metrics.

**Birds Eye View:** Zou and Li (2021) [95] observed that recent works introduced urban-level datasets acquired using UAVs and include both images and 3D point clouds. In general, the created photogrammetric 3D point clouds present differences to those acquired using LiDAR sensors, for example the existence of RGB values along with the 3D geometric information. As such, they compared the category overlap of the points along the Z-axis in the photogrammetric and LiDAR point clouds and concluded that most overlapping points had the same class as the top one. Hence, they proposed the Bird's-Eye View (BEV) projection as the more suitable one for the 3DSS of the photogrammetric point clouds created using UAVs. To be more specific, the creation of the BEV images was conducted by using a sliding window process over the 3D points. Firstly, the scale, size and moving step hyperparameters of the sliding window were empirically defined after experiments. Then, the points in each sliding window were sorted according to their x and y coordinates to find the minimum and maximum values. Finally, each sliding window created a BEV image which included the RGB and altitude information. Afterwards, a sparse BEV image completion process was applied using 2D max pooling operation to improve the sparse information of the projected points onto the XY plane, resulting to the final BEV images. Furthermore, Zou and Li (2021) proposed a multimodal 2D segmentation UNet, which exploits both the RGB and the altitude modalities. Additionally, the authors stated that the RGB values played a significant role in the segmentation performance. To conclude, the authors evaluated their approach over different SotA methods (test set) using the validation set of the Sensat-Urban [96] benchmark.

**Multiple Projections:** Alnaggar et al. (2020) [97] stated that both the BEV and the spherical projection images provide useful features for 3DSS, hence they can complement each other to decrease the loss of information due to the projection of 3D point clouds onto 2D space. The spherical projection image representation is similar to the RangeNet++ [87]. The BEV representation was a projection onto the XY plane and discretization based on a 2D grid with specific dimensions, while the channels were similar to the spherical image. Moreover, they proposed a two-branch network, one for each projection, providing two different predictions which were finally fused by adding them. To be more specific, the spherical branch was an encoder-decoder architecture using the MobileNetV2 [98] backbone, while the BEV branch was based on a UNet [99] architecture. Before fusion, the segmented images were inserted into a post-processing step. Specifically, the 3D points were projected onto each segmented image. Then, the neighborhood of each point was defined using a 2D square window around the projected point. Afterwards, a score vector for each point was calculated using a weighted sum of the SoftMax probabilities of all the pixels in the neighborhood. The weights were calculated based on the distance between the 3D point under investigation and those represented by the pixels in the neighborhood. More concretely, the authors gave attention to the nearest points rather than the distant ones by defining different weight values based on their distance. The final score vectors, one for each projection, were defined into the fusion step. To conclude, the authors provided an evaluation of the proposed approach in addition to an experimentation using data augmentation techniques.

Qiu, Yu and Tao (2022) [2] observed that there were two widely used, complementary projections by the dimensionality reduction methods for 3DSS: the spherical range images (RV) and the top down images i.e., BEV. Additionally, they mention that the methods which exploited both representations were generally using late fusion of the predicted labels, ignoring the complementary geometric information between the different views. Hence, the authors introduced the Geometric Flow Network (GFNet) aiming to exploit the geometric correspondences of each view, using the original point cloud as a bridge, in an align-before-fuse fashion. The proposed real-time network processed each view separately by a two-branch ResNet [75] based encoder-decoder network, which adopts an ASPP [100] module at the bottleneck. To be more specific, the range view images are created by using an improved spherical projection approach proposed by Triess et al. (2020). Each spherical image has five channels, similar to Milioto et al. (2019). Furthermore, the BEV images were created using the top-down orthogonal projection replacing the cartesian coordinates with relative polar coordinates. Each BEV image had nine channels i.e., three cylindrical distances (x, y, z) to the center of the BEV grid, three cylindrical coordinates (x, y, z), two cartesian coordinates (x, y) and remission. Moreover, the authors proposed the Geometric Flow Module (GFM), which was divided into the Geometric Alignment (GA) and Attention Fusion (AF). In fact, GA included the geometric transformation from the RV to BEV and vice versa for feature fusion. To calculate the transformation matrices, the authors used the original point cloud (PC) as a bridge, i.e., RV to PC to BEV. Furthermore, AF module concatenated the single-view features, e.g., from the RV image, with the transformed features, e.g., BEV to RV, then applied self-attention on the concatenated features to obtain attention scores, and finally combined them with the single-view features by using a residual connection, resulting in the final features. In general, several GFMs were located between the two ResNet decoders, obtaining a feature map for each view. Finally, the 2D predictions of each branch were utilized along with grid sampling and KPConv [73] to find the per-point predictions. In fact, the KPConv layer was exploited instead of the kNN post processing approach presented by Milioto et al. (2019) to create an end-to-end learnable approach.

***Table 4:*** *Mean Intersection over Union (mIoU) and Overall Accuracy (OA) for Different Benchmark Datasets for the Dimensionality Reduction Based Methods*

| Algorithm | Year | Semantic3D | | ScanNet | | SemanticKITTI | | S3DIS | | Sensat Urban | |
|---|---|---|---|---|---|---|---|---|---|---|---|
| | | mIoU | OA | mIoU | OA | mIoU | OA | mIoU | OA | mIoU | OA |
| TangentConv | 2018 | 66.4 | 89.3 | 43.8 | 55.1 | 40.9 | --- | 52.8 | 82.5 | 33.30 | 76.97 |
| SqueezeSeg | 2017 | --- | --- | --- | --- | 30.8 | --- | --- | --- | --- | --- |
| SqueezeSegV2 | 2018 | --- | --- | --- | --- | 39.7 | --- | --- | --- | --- | --- |
| RangeNet++ | 2019 | --- | --- | --- | --- | 52.2 | 89.0 | --- | --- | --- | --- |
| SqueezeSegV3 | 2020 | --- | --- | --- | --- | 55.9 | 89.5 | --- | --- | --- | --- |
| SalsaNext | 2020 | --- | --- | --- | --- | 59.5 | 90.0 | --- | --- | --- | --- |
| FPS-Net | 2021 | --- | --- | --- | --- | 57.1 | --- | --- | --- | --- | --- |
| MINet | 2021 | --- | --- | --- | --- | 55.2 | --- | --- | --- | --- | --- |
| Efficient BEV | 2021 | --- | --- | --- | --- | --- | --- | --- | --- | --- | 91.37 |
| MPF | 2020 | --- | --- | --- | --- | 55.5 | --- | --- | --- | --- | --- |
| GFNet | 2022 | --- | --- | --- | --- | 65.4 | 92.4 | --- | --- | --- | --- |

### 5.3. Discretization Based Methods

Riegler, Ulusoy and Geiger (2017) [102] observed that the SotA discretization-based methods required exhaustive dense convolution operations, i.e., they took into account the 3D empty space, ensuing slow computations on the downstream applications like 3DSS. In this regard, Riegler, Ulusoy and Geiger (2017) proposed the OctNet, a 3D convolution-based network which avoided the empty space, proposing a new intermediate sparse representation of the 3D data. To be more specific, the new representation of the 3D data was created by subdividing the high-resolution 3D data hierarchically into octrees, taking into account the density of the 3D points and stopping when achieving the predefined resolution. Moreover, the authors proposed the Hybrid Grid-Octree Data Structure (HGODS), by stacking several shallow octrees similar to Miller, Jain and Mundy's (2011) [103] representation. The scope of the HGODS was the creation of a sparse representation which permitted rapid data access using bit strings, because several downstream applications like 3DSS commonly require the definition of points' neighborhoods, e.g., for convolution or pooling. To this end, the authors stated that the discretization of the data followed their density, avoided the empty space, concentrated the computations only on the non-empty regions, and improved the computational and memory requirements. Furthermore, the authors presented the basic OctNet network operations i.e., Convolution, Pooling and Unpooling, along with different applications like 3D classification and 3DSS. The 3DSS OctNet was an encoder-decoder UNet-shaped network. To conclude, the OctNet network was trained using different voxel sizes and features (RGB, normal vector, binary voxel occupancy, height above ground) in combination with data augmentation.

Su et al. (2018) [88] proposed the SPLATNet architecture which exploited high-dimensional lattice space to perform the convolution operation on a set of features. To be more specific, the authors stated that the SotA 3DSS methods usually transformed the raw 3D point clouds into 2D - 3D grid representations to use the convolution operation on them. However, those transformations resulted in a loss of information, and thus the authors proposed the high-dimensional lattice space for the convolution operation aiming to reduce the loss of information of the SotA methods. Hence, they proposed the SPLATNet architecture stepping on the Bilateral Convolution Layers (BCLs) introduced by Miller, Jain and Mundy (2011). In fact, BCL can easily be operated into high dimensional lattice spaces, e.g., the 6-dimensional filtering space XYZRGB, which was a strong property in order to be chosen as the main operation of the SPLATNet architecture. More concretely, BCL was incorporated into the Splat, Convolve and Slice steps, each of which was written as matrix multiplication. Firstly, the Splat operation projected the input features into the permutohedral lattice space defined by the lattice features, using the barycentric interpolation, along with a scale factor. Secondly, the Convolve operation was applied onto the lattice space, defining N-dimensional filter weights similar to the conventional convolution operation. Finally, the Slice operation was the opposite of the Splat operation, but with the opportunity to choose if the resulted point cloud will be the same or different from the input one. The authors included an in-depth analysis of BCL advantageous properties regarding the operation using 3D point clouds. Furthermore, they performed 3DSS using two variations of SPLATNet, one using only 3D point clouds and one including images in addition to the 3D point cloud data. A unique characteristic of lattice space was the scale, which was strongly associated with the receptive field of the convolution operation. The authors included a detailed analysis regarding the lattice scale factor. To conclude, SPLATNet architecture was evaluated on a series of downstream applications like 3DSS and 3D part segmentation.

Choy, Gwak and Savarese (2019) [104] proposed a 4D spatio-temporal convolutional neural network to interpret 3D video scan data, i.e., 3D scenes spanning over different timestamps. In general, the 3D data like point clouds or voxels included empty areas which should be avoided during the calculations, because they did not contribute to the performance of the networks. The authors constructed two N-dimensional sparse tensors, instead of the 3D data representation, to avoid the empty areas of the data. To be more specific, the sparse tensors were a more useful and homogeneous representation of the data, especially for high dimensional spaces. Additionally, the authors included a detailed explanation of the generalized sparse convolution which had multiple advantages like it was efficient, it could be applied to high dimensional spaces, and it could be used to reproduce the milestone 2D techniques into high dimensional space. The aforementioned units, e.g., sparse tensors and the generalized sparse convolution, were included in the open-sourced Minkowski Engine. More concretely, the Minkowski Engine included the Sparse Tensor Quantization, Generalized Sparse Convolution, Max Pooling, Global / Average Pooling, Sum Pooling and Non-spatial Functions algorithms, which were presented in detail, including a pseudo-code explanation for each of them. Using the components of the Minkowski Engine, the authors proposed the Minkowski Convolutional Neural Network, using ResNet [75] or UNet [99] as the backbone architecture. In general, spatio-temporal convolutions had two problems. The first one was that the computational cost was exponentially increased along with the dimensions. The second problem was that the predictions were not consistent between the different timestamps. Hence, the authors investigated the exploitation of non-conventional kernel shapes to overcome the computational cost problem, and the use of trilateral stationary conditional random fields (CRFs) for the second problem. To be more specific, the authors investigated the use of different kernels between the spatial and temporal dimensions resulting in a hybrid-shaped kernel which outperformed its counterparts i.e., the tesseract kernels. Furthermore, the proposed CRF contained a stationary 7D kernel (3D space, 1D time and 3D color space), and thus it was called Trilateral Stationary CRF. To conclude, the Minkowski Network was evaluated in many downstream tasks, including 3DSS, using different metrics and benchmark datasets.

***Table 5:*** *Mean Intersection over Union (mIoU) and Overall Accuracy (OA) for Different Benchmark Datasets for the Discretization Based Methods*

| Algorithm | Year | Semantic3D | | ScanNet | | SemanticKITTI | | S3DIS | |
|---|---|---|---|---|---|---|---|---|---|
| | | mIoU | OA | mIoU | OA | mIoU | OA | mIoU | OA |
| OctNet | 2017 | 50.7 | 80.7 | 18.1 | 76.6 | --- | --- | 26.3 | 68.9 |
| SPLATNet | 2018 | --- | --- | 39.3 | --- | 22.8 | --- | --- | --- |
| MinkowskiNet | 2019 | --- | --- | 73.6 | --- | 54.3 | --- | 65.4 | --- |

### 5.4. Graph Based Methods

Wang *et al.* (2019) [105] stated that the convolutional neural networks (CNNs) achieved high end results in many 2D downstream tasks. However, using the convolution operation in 3D space was not a straightforward process. Hence, they proposed the EdgeConv, an operation similar to the traditional convolution, for 3D downstream applications like classification and semantic segmentation. Additionally, they proposed the DGCNN architecture, which was based on the PointNet architecture but without feature transformations and including the EdgeConv operation. Specifically, the proposed architecture was based on the Point cloud Transform Block and the EdgeConv Block. The former defined a 3x3 point cloud transformation, by concatenating their global and local coordinates, to align it into a canonical space. The local coordinates were defined subtracting the kNN neighboring point coordinates to the center point coordinates. The latter, i.e., the EdgeConv operation, applied a convolution based operation on graph edges, similar to the graph neural networks. To be more specific, the authors built a local neighborhood graph for each point, defining their neighborhood i.e., connecting the neighboring points with edges. More concretely, the EdgeConv operation was defined using (i) a nonlinear and (ii) an aggregation function. Firstly, the local structure of each point in the given point cloud was defined by computing a directed graph e.g., kNN graph. In fact, the computed graph had the 3D points as nodes connected to their neighboring nodes with edges. Secondly, the calculated edges were enriched with features using the predefined nonlinear function i.e., edge features. Finally, the output of each EdgeConv layer was calculated applying the aggregation function on the previously calculated features. In general, the output point cloud had the same number of 3D points but with more edge features than the input point cloud. Furthermore, the authors described in detail, the choice of different nonlinear and aggregation functions and how some SotA methods like PointNet, can be considered as a subset of the EdgeConv operation i.e., by defining the appropriate set of the nonlinear and the aggregation functions. Finally, they introduced the asymmetric edge function which was selected for the DGCNN architecture because they stated that it was thoroughly combined the global and local point cloud features and also that can be defined as a shared MLP. Moreover, the authors found that finding the graph kNN in feature space other than the Euclidean space, was beneficial for their network. Thus, they proposed to update the graph kNN dynamically into each layer. To conclude, the authors presented an evaluation of the DGCNN architecture in a series of high level tasks like 3D classification, 3D part segmentation and indoor 3DSS, using several SotA methods as benchmark.

G. Li *et al.* (2021) [106] observed that the SotA GCN models were shallow, basically due to the vanishing gradients problem and the high complexity in the computation of backpropagation. Hence, they investigated the transfer of several techniques like residual/dense connections and dilated convolutions from CNNs to GCNs to create deep GCNs for different high level tasks like 3DSS. First and foremost, the authors defined the meaning of graphs. Furthermore, they defined the general idea of GCNs i.e., the definition and application of the update function (MLPs, Gated Networks etc.) or the non-linear function [105] and the aggregate (mean, max, attention etc.) function. In their framework, the authors chose the max aggregation function and an MLP with batch normalization and ReLU as the update function. Moreover they described the differences between the fixed graph representation, commonly used in GCNs, with the Dynamic Graph representation which was presented by Wang *et al.* (2019), concluding that the later representation was more beneficial for the networks. Hence, they included a Dilated kNN function to dynamically change the points neighbors in each layer to increase the GCNs receptive field. Afterwards the author introduced in detail the *Residual Connections for GCNs* and the *Dilated Aggregation for GCNs* operations. Then they analyzed different *Deep GCN variants* e.g., they included the introduced operations into the EdgeConv one, to train deeper GCNs etc. Most importantly, the authors included an detailed experimental process using a plethora of Graph Learning techniques. Also, they proposed three variations of the GCNs backbones the *PlainGCN* the *ResGCN* and the *DenseGCN*. The selected backbone was the only different part among the proposed architectures during the experiments i.e., the under investigation part of the evaluation process was the backbone architecture. More concretely, the *PlainGCN* architecture was similar to DGCNN. In addition to the PlainGCN architecture, the dynamic dilated kNN graph and the residual graph connections were added to form the *ResGCN* architecture. Finally, the DenseGCN was defined by changing the residual with dense graph connections. To conclude, the authors presented an in-depth comparison among the different architectures and modules using many ablation studies.

***Table 6:*** *Mean Intersection over Union (mIoU) and Overall Accuracy (OA) for the S3DIS Benchmark Datasets for the Graph Based Methods*

| Algorithm | Year | S3DIS | |
|---|---|---|---|
| | | mIoU | OA |
| DGCNN | 2019 | 58.2 | 84.1 |
| Deep GCN | 2021 | 60.0 | 85.9 |

## 5.5. Hybrid Methods

**Discretization, Point & Dimensionality Reduction Based Methods:** Zhang *et al.* (2020) [107] introduced the PolarNet architecture aiming to cope with the LiDAR point clouds irregularity property and the use of many detailed semantic classes while retaining a real time perception performance. The authors described the very important role played by the size of the receptive field to 2DSS performance. However, they concluded that in 3DSS the shape of the receptive field was also important in addition to the size. Firstly, they proposed a bird's eye view (BEV) representation observing that this view of the LiDAR scans organizes the points in rings with different radius. Moreover, they described that a partition of the BEV representation using a Cartesian grid will result to an uneven distribution of points into each grid cell. To be more specific, the grid cells that were closer to the LiDAR sensor will contain more points while those that were further away will contain fewer points. Additionally, there were many empty cartesian grid cells. Hence, they proposed a polar grid feature learning approach, instead of the cartesian grid, to exploit the rings of different radius described earlier. More concretely, the authors first calculated the azimuth and radius of each point onto XY plane, using as the origin the LiDAR sensor. Then, the points were assigned to a grid cell using the calculated azimuth and radius values resulting in the BEV grid cell representation of the input 3D point cloud. In fact, each grid cell was similar to point pillars representation [108] i.e., XY plane coordinates and unlimited spatial extent of the Z direction. Afterwards, the points inside each cell were passed into a kNN-free PointNet followed by a max-pooling operation resulting to a set of fixed length features (1x512). The output features were assigned into a ring matrix taking into account the spatial location of their corresponding grid cell. Moreover, the authors proposed the ring convolution aiming to predict the 3D points labels using the ring matrix as an input. Specifically, the ring convolution was operated on the ring matrix along the radius axis. The authors mentioned that any 2DCNN network can process the created representation by replacing the traditional CNN operation with the discrete ring convolution. Finally, they reshape the ring predictions into a 4D matrix to exploit a voxel-based segmentation loss. To conclude, PolarNet used various techniques inspired by Point, Dimensionality Reduction and Discretization based methods. Also, the authors investigated in depth the performance of PolarNet including several experiments with well-known benchmark datasets and comparisons with many baseline networks.

Gerdzhev *et al.* (2020) [109] proposed the TORNADO-Net architecture which included techniques proposed into the PolarNet [107] and SalsaNext [91] architecture. To be more specific, they used the general idea of PolarNet but they combined features from both BEV and Range images, under the pillar projection learning scheme. Furthermore, similar to SalsaNext, they used the same Range Image creation process as well as a similar to the *Context Module* called *Diamond Contextual Block* which included different 2D techniques to improve the SalsaNet architecture. However, the main contribution of this architecture was the implementation of a total loss function exploiting the combination of weighted cross entropy loss and Lovasz SoftMax loss, proposed in SalsaNext but with an addition of the total variation loss along with different weights for each part. Finally, a cut-off thresholding [87] and a post processing approach using kNN [87] were exploited to improve the networks performance. To conclude, the authors presented several quantitative and qualitative analysis achieving high end results, evaluated measuring the mIoU on the SemanticKITTI benchmark.

Liu et. al. (2023) [110] stated that the cross-modal and cross-view fusion approaches had not been thoroughly investigated yet, while the information gathered using multi-modal data and different views of the point clouds were complementary to each other. Furthermore, the authors observed that the 3D point clouds carried detailed geometric information while the images carried detailed semantic information. Hence, they proposed the UniSeg architecture which fused the features gathered from the -point, -range and -voxel representations of the 3D point cloud in addition to those gathered from 2D images to improve the performance of both 3DSS and 3D panoptic segmentation. Firstly, the authors extracted the -point, -range, -voxel and image features using conventional methods i.e., MLPs, spherical projected images, max pooling on the voxel representation and the ResNet architecture respectively. Secondly, they proposed, the *Learnable cross-Modal Association (LMA) module* and the *Learnable cross-View Association (LVA) module* to fuse the -voxel and -range features with the image features and to transform the fused features into the point space and then combine them using different views of the point cloud respectively. In more detail, the LMA module fused the features gathered from the voxel and range representations with the image features resulting to the set of the image-enhanced voxel and range view features. To be more specific, for each voxel, the voxel features and their corresponding image features were passed into a multi-head cross attention module resulting in the enhanced voxel features. The range-view features were processed similarly to the voxel features, resulting in the enhanced range features. Before, the enhanced features were fed into the LVA module they were firstly transformed into the point space using the trilinear and bilinear interpolation in order to alleviate the quantity mismatch problem. Afterwards, the enhanced features were fed into the LVA module. Moreover, all the features, i.e., the transformed -range, - voxel features and the point features, were concatenated resulting to the multi-view features. Then, the multiview features were further processed producing the view-wised adapted features, using the original point space features and they were projected back to the voxel and range representation forming the final set of features for the 3DSS. To conclude, the UniSeg architecture was exploited the created features using different heads depending on the application i.e., 3DSS or 3D panoptic segmentation while it was evaluated using different benchmark datasets like SemanticKITTI, nuScenes and Waymo Open.

**Graph & Discretization Based Methods:** Yan *et al.* (2020) [111] observed that the LiDAR 3D point clouds were characterized by sparsity resulting to underperformance on the 3DSS task. Additionally, they stated that combining multi temporal scans could be used to create a dense representation of the scene and thus to improve the performance of 3DSS. However, SotA multi temporal methods, were commonly used only previously, to the current scans and thus they cannot exploit the upcoming frames. Also, they introduced time-consuming feature aggregation techniques like kNN, which harden their application on the self-driving task. Hence, Yan *et al.* (2020) [111] proposed the JS3C-Net architecture which exploited both a 3DSS and 3D Semantic Scene Completion (3DSSC) modules to equally improve them, along the execution. To be more specific, the 3DSS module used a U-Net architecture implemented based on the SparseConv operation introduced by Graham, Engelcke and Maaten (2018) to create the voxel-based 3DSS output. Afterwards, the voxel-based predictions were transferred to the original 3D point cloud by nearest-neighbor interpolation. Then, the transformed features were fed into three MLPs. The first MLP transformed the point-wise features into a shape embedding (SE) which was then fed into the *Point-Voxel Interaction* which was a submodule of the 3DSSC module. The created SE was then used diversly. Firstly, the point-wise features were fed into the second MLP and then were fused with the SE using an element-wise summing operation. The fused features were fed into the third MLP resulting to the 3D point cloud semantic segmentation predictions of the 3DSS module, which was the input to the 3DSSC module. Moreover, the proposed 3DSSC module was first densely voxelized the produced 3DSS map creating a 3D high-resolution volume which was further processed using convolution, pooling, concatenation and upsampling layers along with skip connections to use multi-scale features. Finally, the 3DSSC module output was a set of voxelized coarse completion features which were also fed into the *Point-Voxel Interaction* submodule. Furthermore, the *Point-Voxel Interaction* includes the SE from the 3DSS module and the coarse voxel based completion from the 3DSSC module and aimed to exploit them to transform the coarse completion into a fine one. To achieve that, the coarse completion non-empty voxels centers were first collected to create a new point cloud. Then, the k-nearest neighbors between the SE and the new point cloud were gathered. Finally several GCN layers, inspired by Wang *et al.* (2019), were stacked together to get the refined completion exploiting the kNNs. To conclude, the authors presented an evaluation of their hybrid architecture comparing its performance with SotA methods using data from the SemanticKITTI benchmark.

**Point & Discretization Based Methods:** Liu *et al.* (2019) [113] observed that the limitations of the Point-Based Methods and the Voxel-Based Methods were usually complementary to each other and thus a combination of techniques from both can result to memory efficient 3D deep learning models. On the one hand, the Voxel-Based methods quantized the given point cloud resulting to a loss of information while creating a regular representation with good memory locality i.e., mimicking the 2D grid representation. On the other hand, the Point-Based methods had a significant latency in order to overcome the irregular memory access as well as to find the relative distances among the point neighbors and the center point. Basically, due to the irregular property of point clouds but with the advantage of a small memory footprint. To this end, the authors proposed the PVCNN architecture as a hybrid approach which exploited the advantages of Voxel and Point-Based methods using the proposed Point-Voxels Convolution (PVConv) operation. To be more specific, the PVConv operation was decomposed into two branches, the *Voxel-Based Feature Aggregation (VBFA)* and the *Point-Based Feature Transformation (PBFT)*, to capture coarse-grained features and fine-grained features respectively. More concretely, VBFA first normalized the given point cloud before the voxelization. To achieve that, the 3D points were expressed in respect to the gravity center of the point cloud and divided by the largest distance. Finally, the points were scaled and translated in order to span between 0 and 1. The normalized point cloud was then fed into the voxelization step to create the 3D volumetric representation. In this regard, the authors presented an informative figure comparing the GPU memory requirements, the information loss and the voxel resolution. Moreover, multiple 3D convolutions were applied using the 3D voxel grid into the Feature Aggregation step, followed by batch normalization and nonlinear activation function. Finally, the authors exploited the trilinear interpolation to map the voxel-based features to the point cloud domain. Furthermore, the PBFT was applied to an MLP directly on the original point cloud resulting to a set of point-wise features. Finally, the coarse-grained and the fine-grained features, created from the VBFA and the PBFT respectively, were fused using the addition operation. To conclude, the authors included thorough experimentation and evaluation of the proposed architecture in respect to different SotA methods stating that PVCNN and PVConv, were efficient and effective.

Tang *et al.* (2020) [114] observed that the PVConv [113] and the Sparse Convolution [104] operations struggled to capture the small instances of a 3D scene, especially using hardware with limited memory, due to the point cloud coarse voxelization and the aggressive downsampling, respectively. To overcome these limitations, the authors proposed the *Sparse Point-Voxel Convolution (SPVConv)* module along with a *3D Neural Architecture Search (3D-NAS)* process. Specifically, SPVConv operation had two branches, the point based and the voxel based branch, which communicated through sparse voxelization and devoxelization operations. The former preserved the high resolution details while the latter operated on a multi receptive field manner. Firstly, the original point cloud was sparsely voxelized by exploiting a GPU based hash table representation of the data. Furthermore, the hash table was exploited into both the devoxelization and feature aggregation procedures. More concretely, the feature aggregation was implemented using Sparse Convolution residual layers. The information gathered through the voxel based branch was transformed back into the 3D point cloud through the devoxelization operation. In parallel, an MLP was applied on the original point cloud resulting to a fine detailed set of features. Both features, i.e., from the voxel based and point based branches, were finally fused to be used for point labelling. Moreover, the authors proposed the 3D-NAS process in which multiple models were automatically assessed in order to automatically find the most efficient one for each application. To this end, the authors defined the searching space by incorporating the channel number and the network depth and finally using an evolutionary architecture search regarding a set of predefined hardware constraints to find the most efficient model. To conclude, the best model was evaluated on 3DSS among others downstream tasks, using different SotA models and the SemanticKITTI dataset.

Rosu *et al.* (2020) [115] presented the LatticeNet a hybrid network which exploited both the PointNet architecture and lattices, for 3DSS. Firstly, the authors described in detail the notation used along the paper, the permutohedral lattice structure as well as the existing lattice operations i.e., *Splatting*, Convolving and *Slicing*, similar to the SplatNet architecture. However, they observed that the weights used into the *Splatting* and *Slicing* operations, from SotA methods, were defined using the barycentric interpolation and stated that a better interpolation could be defined by changing the weights into a set of learnable parameters. Furthermore, they proposed four new lattice operations, the *Distribute*, *Downsampling*, *Upsampling* and *DeformSlicing*. Firstly, *Distribute* operation aimed to enrich the lattice vertices with a list of features. To achieve that, the coordinates and the features of the contributing points were concatenated and processed using PointNet resulting to the final lattice vertices values. The coordinates were defined in respect to the mean value of the contributing points, before the distribution operation, to include the local information of the semantic class. Additionally, the authors stated that the proposed operation aimed to avoid the naive summation of *Splatting* operation and thus to preserve the information through the network. The Downsampling and Upsampling operations followed the same idea. Firstly, a coarse lattice was produced by repeatedly dividing the 3D point cloud coordinates by two. In each repetition, a coarse lattice was produced while the previous lattice was referred to as the finer lattice. On the one hand, during the *Downsampling* operation, the coarse lattice was embedded into the corresponding finer lattice by scaling it up by two, resulting to a set of coarse vertices into the finer vertices space. Then the convolution operation was applied on the finer lattice vertices, using a step equal to one, to get the coarse vertices values, an idea similar to the strided convolution. On the other hand, the *Upsampling* operation embedded the finer vertices into the coarse vertices space by dividing them by two and then the convolution operation was applied using a step equal to minus zero point five, similar to the transposed convolution. Moreover, the *DeformSlicing* operation aimed to improve the Slicing operation by learning to shift the position of the barycentric coordinates i.e., allowing a data-driven interpolation instead of a simple barycentric interpolation. Finally, the authors proposed a U-Net structure network, including firstly, a *Distribute* operation, then a series of ResNet blocks, *Downsampling* and *Upsampling* operation and finally a *DeformSlicing* operation. To conclude, the authors presented several experiments and ablation studies to evaluate the performance of the LatticeNet architecture.

Cheng *et al.* (2021) [116] stated that the SotA methods for 3DSS were characterized by high computational complexity and inability to gather the fine details of small objects. To overcome these limitations, they proposed the $(AF)^2$-S3Net architecture, exploiting techniques from both the discretization and point based methods. To be more specific, the proposed network used the MinkNet42 [104] as backbone, enriched with the *Attentive Feature Fusion Module (AF2M)* and the *Adaptive Feature Selection Module (AFSM)*. First and foremost, the input point cloud was transformed into a sparse tensor, similar to the MinkowskiNet approach, containing the 3D cartesian coordinates along with per point normals and intensity, as features. Afterwards, the sparse tensor was fed into the hybrid AF2 module in which multi-scale point-wise and voxel based features were extracted using different branches. In total, three branches were used, the first one was focused on the fine details of smaller objects while the other two on global features using attention maps. Finally, the multi-branch features were fused using summation. Afterwards, the fused features were further processed by convolutional layers, in an encoder-decoder fashion, while each branch features were fed independently into the AFS module. In the AFS module, each branch features were firstly processed using convolutional layers resulting to a new set of features for each branch. Then, the new features were fused and fed into a shared squeeze re-weighting network [117] resulting to the output of the module which was passed to the last transposed convolution for learning stability purposes. Finally, the predicted labels firstly for the sparse tensor and then for the original point cloud were exported from the network decoder. To conclude, the authors presented an evaluation of the proposed approach on two benchmark datasets for 3DSS and a qualitative analysis using several SotA methods.

Zhu *et al.* (2021) [118] stated that the dimensionality reduction based methods for 3DSS of outdoor scenes, inevitably lose a significant amount of information due to the projection operation. Furthermore, they stated that the discretization based methods were slightly improved the dimensionality reduction methods. Moreover, they observed that the outdoor 3D point clouds suffer from sparsity and varying density. To overcome these limitations, the authors proposed a new framework which incorporated two components the *Cylindrical Partition (CP)* and the *Asymmetrical 3D Convolution Network (A3DCN)*. To be more specific, the uniform cube voxelization process did not take into account the varying density of LiDAR outdoor point clouds i.e., the cell size was independent to the distance. The general idea of the authors was to use the cylindrical partition to follow the point density i.e., the cell size to be increased following the distance. Hence, the farther cells contained more points than their counterparts on uniform cube voxelization i.e., the cells had a more balanced point distribution. To prove that the authors explained the pros and cons of uniform and cylindrical voxelization. More concretely, the CP component first transformed the 3D points from the Cartesian to Cylinder coordinate system. Then, the 3D partition was performed. Meanwhile, the original point cloud was fed into multiple MLPs resulting in a set of features which were assigned to their corresponding cylindrical cells gathering the cylindrical features sets. Afterwards, the cylinder was unrolled resulting to the representation which was fed into the A3DCN component. The authors observed that the objects typically occupied the crisscross area of the cylindrical partition and hence these cells should be strengthened compared to the others resulting to an asymmetrical cell operation. First and foremost, the authors defined the asymmetrical upsample and downsample operations. Then the Asymmetrical Residual Block enhances the kernels which operated on the crisscross area and applied a series of asymmetrical downsample and upsample operations. Afterwards, the *Dimension Decomposition based Context Modeling (DDCM)* component was applied to gather the global context of the point clouds by stacking several low-rank features, resulting to a discretization based 3DSS i.e., a label for each cell, which suffered from information loss. To this end, the authors proposed the *Point-wise Refinement Module (PRM)* to enhance the 3DSS output with fine grained details. To be more specific, RPM firstly, projects the 3D convolution features gathered from the aforementioned process to the point-wise features. Finally, a series of MLPS, which communicate with the first set of MLPs were used to fuse point-wise and voxel-wise features resulting to the refined output. To conclude, the authors evaluated their framework on the SemanticKITTI and nuScenes datasets and among different SotA methods.

Yan *et al.* (2022) [119] stated that modality specific 3DSS suffers from the limitations inherited by each sensor capability. Additionally, observed that the LiDAR and images complement each other and thus overcoming some of the limitations of each sensor and hence, fusion-base methods seem to be beneficial for 3DSS. However, different fusion techniques like the point-pixel correspondence construction along with the different Field-of-View (FOV) of the sensors and the more computational resources required, downgrade the SotA fusion based methods. To this end the authors proposed the 2D Priors Assisted Semantic Segmentation (2DPASS) framework aiming to overcome the aforementioned issues. To be more specific, each modality was processed using a modal-specific architecture. Concretely, the 2D encoder-decoder architecture was based on the ResNet34 architecture and the fully convolutional layer (FCN) while the 3D encoder-decoder architecture was based on the Sparse Convolution similar to the Tang *et al.* (2020) implementation, resulting to a 2D and a 3D set of multi scale features. Moreover, the authors aim to use complementary information and thus to create point-pixel correspondences. To achieve that, the authors exploited the perspective projection along with a timestamp calibration between the two modalities. In fact, the authors used the 2D branch only during training using as 2D ground truths the projected ground truth 3D labels. Afterwards, the 2D-3D features were fused using the key component called *Multi-Scale Fusion-to-Single Knowledge Distillation (MSFSKD)*. More specifically, the 3D features were fed into an MLP resulting to a new set of features which were further concatenated with the 2D features through another MLP gathering the fused features. However, similar to the ResNet idea the authors enhanced the original 3D features with the new set of features preserving the modality specific information through the process. Finally, both the fused and the 3D features were fed into independent classifiers gathering the semantic scores. To conclude, the authors evaluated their method on the SemanticKITTI and nuScenes benchmarks among a wide range of SotA methods.

**Point & Graph Based Methods:** Landrieu and Simonovsky, (2018) [120] observed that the SotA methods struggled to process large scale point clouds especially for 3DSS. Hence, they proposed a hybrid architecture which combined the point cloud graph representation with the PointNet architecture. Specifically, the main contribution was the SPGraph representation which was proposed to handle large scale point clouds. To be more specific, the proposed approach divided the 3DSS into four steps, *Geometrically Homogeneous Partition (GHP)*, *Superpoint Graph Construction (SGC)*, *Superpoint Embedding (SE)* and *Contextual Segmentation (CS)*. The first step i.e., GHP, took as input the entire point cloud and aimed to decomposed it into different geometrically homogeneous parts. This process was similar to the simple segmentation. To achieve that, the neighborhood of each point was defined. Furthermore, several neighborhood characteristics were calculated e.g., linearity, planarity, scattering, verticality, elevation etc. based on the covariance matrix of points neighborhood which constitute the features of the points in addition to the observations i.e., color, intensity etc. Finally, an approximate solution of the generalized minimal partition problem, was found using an adjacency graph technique and the $l_0$-cut pursuit algorithm [121] resulting to several point components each of which constitute the homogenous simple shapes called super points. Moreover, the SGC was aimed at create the oriented attributed graph representation called SPG. More concretely, the SPG was a graph with nodes the superpoints i.e., set of 3D points define simple shapes, which were created into the previous step, and edges the adjacency between them, called super edges. The adjacent superpoints were defined using a symmetric Voronoi adjacency graph of the original 3D point cloud while the edge features were defined calculating the covariance matrix of each set of points and finding different features like surface ratio, volume ratio and length ratio etc. Thus, the SPG representation was created. Following, the SE step aimed to find a descriptor for each superpoint. To this end the PointNet architecture was applied to each superpoint which had a reliable number of points and after a rescaling process. However, the original superpoint was included as a feature, after the maxpooling operation to preserve the original shape of the partition. Finally, each point in the superpoints get its label using a Filter Generating Network (FGN). To be more specific, the FGN was based on the ideas of the *Gated Graph Neural Networks*, the *Edge Conditioned Convolutions* and the *Gated Recurrent Units (GRU)* resulting to the point labels. To conclude, the authors included an in depth analysis for each terminology included to the paper along with an evaluation of the proposed algorithm in comparison to SotA methods and using benchmark datasets.

**Dimensionality Reduction & Point Based Methods:** Kochanov, Nejadasl and Booij, (2020) [122] proposed the KPRNet architecture, which exploited a dimensionality reduction and point based method. Firstly, a 2DSS was performed resulting to 2D labels. Then, the 2D labels were projected back into 3D space. The authors observed that a wide variety of methods explored a post processing step to refine the reprojected labels using either a kNN or CRF based approaches. Instead, they proposed using a KPConv layer to predict the final 3D labels. Additionally, the authors stated that most of the dimensionality reduction methods exploited the spherical projection to create the range images. However, they proposed to unfold the scans in a similar way that the LiDAR acquired the data, resulting to smoother range image than the spherical projected range images. Finally, the authors evaluated the KPRNet architecture on the SemanticKITTI dataset and compared the metrics with those of some of the SotA methods.

Alonso *et al.* (2021) [123] stated that the point based methods were computationally expensive due to the 3D point neighborhood search and that the 3D space operations were more complex than those in 2D space. Thus, they proposed the 3D-MiniNet architecture to overcome the aforementioned limitations. Specifically, the 3D-MiniNet architecture was decomposed into three submodules the *Fast 3D Point Neighbor Search (F3PNS)*, *3D-MiniNet* and *Post-Processing*. Firstly, the F3PNS module received the original 3D point cloud and projected it using the spherical projection. Then, the neighborhood of each point was gathered using a sliding window onto the spherical image. Afterwards, the features of each point of a neighborhood, were augmented computing the relative features to the mean point of them, resulting to a set of eleven features for each point. Furthermore, the 3D-MiniNet module had two submodules the *Projection Learning Module (PLM)* and the *2D Segmentation Module (2DSM)*. Firstly, the PLM extracted three types of features, the Local, Context and Spatial and finally fused them. To be more specific, the input group of points, defined into the F3PNS, were fed into the *Local Feature Extractor (LFE)* extracting PointNet like features. Meanwhile, after the second layer of the LFE the points and their features were fed into the *Context Feature Extractor (CFE)* in which the neighbor search implementation was applied using different window sizes resulting to an informative set of features. Both features, i.e., from LFE and CFE were concatenated and fed into the max pooling operation. Additionally, the input groups were also fed into the *Spatial Feature Extractor (SFE)* in which a convolution based implementation was applied. Then, the fused features were further concatenated with the SFE features and fed into the Feature Fusion (FF) submodule which fused the concatenated features using a three steps process. Firstly, the concatenated features were fed into a reshaping process. Meanwhile, the concatenated features were processed using average pooling, a 1x1 convolution and a sigmoid function. Afterwards self-attention was applied by multiplying the reshaped features and the computed features. Finally, the features were reduced to fed into the 2DSM module. More concretely, the 2DSM contained two branches. The first branch took as input the output of the PLM and processed using the MiniNet architecture. The second branch took as input the spherical image and processed it to extract high resolution features. Finally, the 2DSS labels were extracted and reprojected back into 3D space.

Afterwards a kNN based postprocessing approach similar to Milioto *et al.* (2019) was applied. To conclude, the authors evaluated the 3DMiniNet architecture using the SemanticKITTI dataset and compared it with different SotA methods.

Robert, Vallet and Landrieu (2022) [124] proposed a hybrid 3DSS approach which exploited 3D point clouds and a set of images along with their poses. The authors stated that, in general, the images were better capturing the textural and contextual information than the 3D point clouds do. Hence, the proposed method aimed to assess the given images based on their viewing conditions and finally use them in combination with the 3D point cloud, for 3DSS. First and foremost, a *Point-Pixel Mapping (PPM)* process was constructed to link each 3D point with a set of images in which it was visible. To achieve that, a similar to Z-buffering method was created resulting to a set of images for each 3D point. Afterwards, its image-point pair was assessed to extract the viewing conditions using a vector which included the normalized depth, some local geometric descriptors, the local density and the viewing angle in respect to the normal vector, among other metrics. Furthermore, the authors exploited the input data along with the PPM outputs to gather diverse features for each point. The authors stated that based on the viewing conditions found earlier, each image could be used diversly contributing to different types of information like detailed textural information or important contextual cues etc. To this end, the authors exploited a deep set architecture to predict a vector which represented the quality of the images correspond to a 3D point. Moreover, the predicted quality vector was fed into the SoftMax function to get the attention scores. In the case that the images' quality was poor the authors blocked the extraction of relevant features from them and relied only on the geometric information, using a gating parameter. Finally, the features gathered for each image in the image set of a 3D point, were fused and assigned to it. The aforementioned described method was called *Multi-View Feature Aggregation Method* by the authors and was exploited under a *Bimodal Point-Image Network* for 3DSS purposes. To be more specific, the proposed bimodal network was composed of a 2D FCN, a 3D encoder-decoder network and a fusion of the 2D-3D features strategy. To conclude, the authors investigated different fusion techniques i.e., early, intermediate and late and evaluated their network on three benchmarks (S3DIS, ScanNet and KITTI-360) against several SotA methods.

Wang, Zhu and Zhang (2022) [125] observed that autonomous vehicles capture sequences of LiDAR data while the SotA 3DSS methods usually processed them using a single frame at a time. Hence, they proposed a range residual spherical image representation, aiming to capture both the spatial and temporal information of LiDAR sequential data for 3DSS. More specifically, the proposed approach called Meta-RangeSeg, was composed by three steps the Range Residual Image Generation, the Feature Extraction and the Post processing. The former step created a nine-channel spherical residual image representation of the sequential 3D point clouds. To be more specific, the authors called the proposed representation as residual because they included three residual channels to the created spherical images, in addition to the remission, the 3D coordinates and an indicator m, which defined if a pixel position was a projected point or not. More concretely, the residual channels number was equal to the sequential frames that were used in addition to the current frame. The authors included three previous LiDAR scans hence the residual channels were three the d1, d2 and d3. Specifically, the residual channels were created applying three steps. Firstly, the 3D point clouds of the previous frames were transformed into the coordinate system of the current frame. After, the transformed point clouds were projected, creating range images. Finally, the residual images were created by calculating the absolute difference among the range images of the previous frames and the range image of the current frame. Moreover, the created spherical images were fed into the feature extraction step, which included the MetaKernel and the U-Net blocks. The MetaKernel block included a sliding 5x5 window on the spherical image, used to indicate the point neighbors. Then, the relative 3D coordinates and range of the neighborhood in respect to the center point were calculated and fed into an MLP. Finally, element-wise product between the learned weights and the MLP features was calculated followed by a concatenation and 1x1 convolution operations, resulting to a set of Meta-features. Afterwards, the meta-Features were fed into a U-Net encoder – decoder network with four downsampling layers and upsampling layers with skip connections. Hence, two types of features were produced the meta-features, using the MetaKernel, and the multi-scale features, using the meta-features fed into a UNet network. Following, the *Post Processing* step was applied, taking as input only the range channel, as the most valuable one regarding the authors, the multi-scale features and the meta features under the *Feature Aggregation Module (FAM)*. Firstly, the range channel was fed into the *Context Module* presented into the SalsaNext [91] architecture, followed by several concatenation, convolution, batch normalization, ReLU and element wise product operations. The resulting features were fused with the multi-scale features creating the Range Guided Features which were concatenated with the Meta-Features. After the concatenation several layers with the same operations i.e., convolution, batch normalization etc. were applied resulting to a set of 2D labels. Finally, the 2D labels were converted into the 3D prediction following the kNN post processing approach presented by Milioto *et al.* (2019). To conclude, the authors evaluated the Meta-RangeSeg algorithm using the SemanticKITTI [67] and SemanticPOSS [126] benchmarks along with the mIoU metric.

**Dimensionality Reduction & Discretization Based Methods:** Wu et. al. (2024) [127] observed that the advancements of 2D deep learning methods were based on the scale principals e.g., dataset size, the number of the model parameters, the size of the receptive field etc. However, the investigation of the scale principals in 3D domain was not a straightforward process due to the limited size and diversity of the available 3D point cloud datasets. Additionally, by observing the advantages of the Sparse Convolution [104] especially using large 3D point clouds, they hypothesized that the 3D deed learning algorithms performance was linked with the scale principals more than the complex architecture designs. Hence, Wu et. al. (2024) [127] motivated by the introduction of scale principals into transformers, proposed the PTv3 architecture. In fact, the SoTA methods commonly investigate a tradeoff between the efficiency and accuracy. To be more specific, the authors emphasize the simplicity and efficiency over the accuracy in order to leverage the scale principals. Firstly, they propose a different to the kNN neighborhood extraction process, by using point cloud serialization along with specific patterns using space filling curves. Specifically, the 3D point cloud was transformed into a structured format using the space-filling curves and transforming the position of every 3D point to an integer which represented its order regarding the space filling curve. During the point cloud serialization, the neighboring points were not changed i.e., the points locality was preserved while a structured format was created. Following, the created representation the authors proposed the *Patch Attention* mechanism which was decomposed into the *Patch Grouping* and *Patch Interaction* designs. In more detail, the former uses the *reordering* and *padding* operations based on the serialization pattern, and neighboring patches respectively. The latter, described the interaction among points belonging to different patches using different techniques like *Shift Patch*, *Shuffle Order* etc. Finally, the overall architecture was designed using a U-Net structure. To conclude, the PTv3 architecture was evaluated on different tasks and benchmarks with remarkable results.

***Table 7:*** *Mean Intersection over Union (mIoU) and Overall Accuracy (OA) for Different Benchmark Datasets for the Hybrid Based Methods*

| Algorithm | Year | Semantic3D | | NuScenes | | SemanticKITTI | | S3DIS | | Sensat Urban | |
|---|---|---|---|---|---|---|---|---|---|---|---|
| | | mIoU | OA | mIoU | OA | mIoU | OA | mIoU | OA | mIoU | OA |
| PolarNet | 2020 | --- | --- | 69.4 | | 54.3 | 90.0 | --- | --- | --- | --- |
| TORNADO-Net | 2020 | --- | --- | --- | --- | 63.1 | 90.7 | --- | --- | --- | --- |
| UniSeg | 2023 | --- | --- | 83.5 | --- | 75.2 | --- | --- | --- | --- | --- |
| JS3C-Net | 2020 | --- | --- | 73.6 | --- | 66.0 | --- | --- | --- | --- | --- |
| PVCNN | 2019 | --- | --- | --- | --- | --- | --- | 58.98 | --- | --- | --- |
| SPVConv | 2020 | --- | --- | 77.4 | --- | 67.0 | --- | --- | --- | --- | --- |
| LatticeNet | 2020 | --- | --- | --- | --- | 52.9 | --- | --- | --- | --- | --- |
| $(AF)^2$-S3Net | 2021 | --- | --- | 78.3 | --- | 69.7 | --- | --- | --- | --- | --- |
| Cylinder3D | 2021 | --- | --- | --- | --- | 67.8 | --- | --- | --- | --- | --- |
| 2DPASS | 2022 | --- | --- | 80.8 | --- | --- | --- | --- | --- | --- | --- |
| SPGraph | 2018 | 76.2 | 94.0 | --- | --- | 20.0 | --- | 63.2 | 86.4 | 37.29 | 85.27 |
| KPRNet | 2020 | --- | --- | --- | --- | 63.1 | --- | --- | --- | --- | --- |
| 3D-MiniNet | 2021 | --- | --- | --- | --- | 55.8 | 89.7 | --- | --- | --- | --- |
| DeepViewAgg | 2022 | --- | --- | --- | --- | --- | --- | 69.5 | --- | --- | --- |
| Meta-RangeSeg | 2022 | --- | --- | --- | --- | 61.0 | --- | --- | --- | --- | --- |
| PTv3 | 2024 | --- | --- | 83.0 | --- | 75.5 | --- | 80.81 | --- | --- | --- |

## 6. LOSS FUNCTIONS IN 3D SEMANTIC SEGMENTATION

The loss or objective function is an integral part of deep learning algorithms. During the research conducted about deep learning algorithms in 3DSS several loss functions have been encountered. For instance, the Boundary loss [125], the Consistency loss [128], the Contextual loss [129], the Contrastive loss [130], [131], the Dice loss [101], the Focal loss [82], [97], [132], [133], [134] and Total Variation loss [109]. However, a detailed analysis of the 3DSS loss functions is out of the scope of this effort. Thus, a small part of them is presented in this section.

**Categorical Cross Entropy Loss:** One of the main losses used in 3DSS is the categorical cross entropy loss, which measures the difference between the predicted and the ground truth value formulated as:

$$L_{ce}(y,\hat{y}) = -\frac{1}{N}\sum_{i}^{N}\sum_{j}^{C} y_{ij}\log\hat{y}_{ij}$$

Where:

- N: Number of samples
- C: Number of Classes
- $y$: Ground Truth Label
- $\hat{y}$: Predicted Label

**Weighted Cross Entropy Loss:** In fact, the cross entropy loss does not take into account the frequency of each class while most of the real world data suffer from the class imbalance problem. Hence the weighted cross entropy loss is frequently adopted in 3DSS as a component of the total loss, to deal with the class imbalance problem using the frequency of each class.

$$L_{wce}(y,\hat{y}) = -\frac{1}{N}\sum_{i}^{N}\sum_{j}^{C}\frac{1}{\sqrt{f_j}} y_{ij}\log\hat{y}_{ij}$$

Where:

- N: Number of samples
- C: Number of Classes
- $f$: Class Frequency
- $y$: Ground Truth Label
- $\hat{y}$: Predicted Label

**Geo Aware Anisotropic Loss:** Liu *et al.* (2020) presented a metric which take into account the variability between the semantic classes of the current voxel with its neighboring voxels. To be more specific, the authors referred to this metric as the Local Geometric Anisotropy which took higher values when a voxel at the edge of the semantic category is under investigation i.e., when the neighboring voxels have a different category than the under investigation voxel. Overall, the LGA is a factor which quantifies the significance of the voxels position and was included to the PA-Loss proposed by the authors. In 3DSS, the LGA is used to recover the fine details of the 3D point cloud and in combination with other losses to form a total loss [116], [136]. Commonly, the LGA is included to the geo aware anisotropic loss formulated as follows:

$$L_{geo}(y,\hat{y}) = -\frac{1}{N}\sum_{i,j,k}\sum_{c=1}^{C}\frac{M_{LGA}}{\Phi} y_{ijk,c}\log\hat{y}_{ijk,c}$$

Where:

- N: Voxel Neighborhood located at I, j, k
- c: Current class of classes C
- $M_{LGA}$: Local Geometric Anisotropy metric
- $y$: Ground Truth Label
- $\hat{y}$: Predicted Label
- $\Phi$: Sliding Window

**Lovasz-SoftMax loss:** In general, the most common evaluation metric in 2D and 3D semantic segmentation is the mean Intersection over Union score (mIoU) or the Jaccard Index. Berman, Triki and Blaschko (2018) proposed the Lovasz-SoftMax Loss to maximize the IoU score. Commonly the Lovasz-SoftMax Loss is included into the 3DSS methods to take into account the fine details presented in the data [91], [93], [94], [109], [125]. However the Lovasz-Softmax loss do not consider the points' neighborhood during the computation which can lead to noise predictions [109]. The Lovasz-SoftMax Loss is formulated as follows:

$$L_{ls} = \frac{1}{|C|}\sum_{c \in C} \overline{\Delta_{Jc}}\left(m(c)\right), \qquad and \;\; m_i = \begin{cases} 1 - x_i(c) & if \quad c = y_i(c) \\ x_{(i)}(c) & otherwise \end{cases}$$

Where:

- $|C|$: Class number
- $\overline{\Delta_{Jc}}$: Lovasz extension of the Jaccard index
- $x_{(i)}(c) \in [0, 1]$ : Predicted Probability of point I for class c
- $y_i(c) \in \{-1, 1\}$: Ground Truth Label

## 7. DISCUSSION

In this section, a fruitful analysis of the examined 3DSS algorithms and datasets, is presented to foster new research directions and applications in the field of 3DSS. First and foremost, a timeline of the included 3DSS methods is presented in Figure 2. Then, the included methods along with their code implementation are summarized in Table 8.

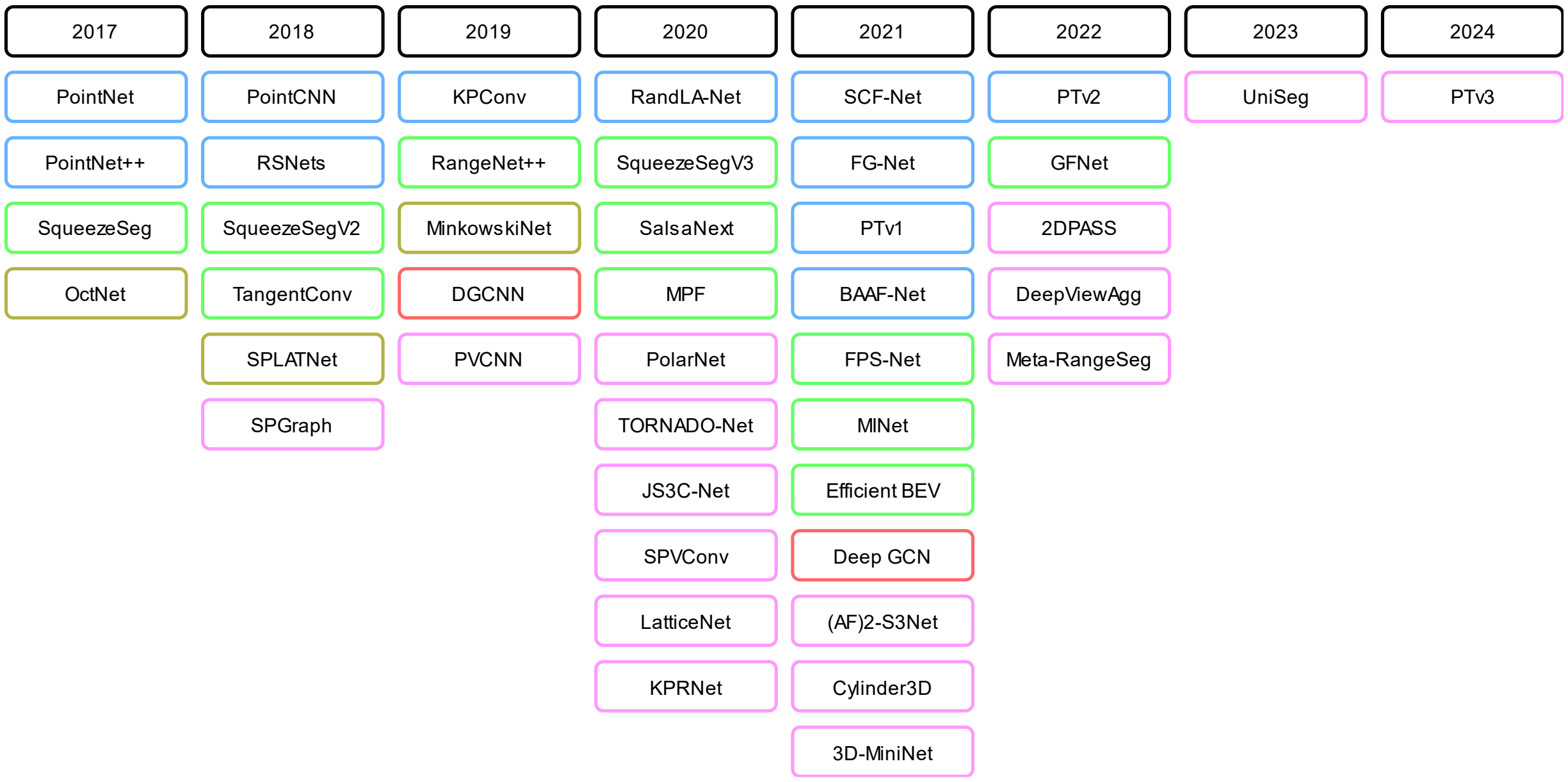

***Figure 2:*** *The timeline of the methods included. Point Based Methods (Blue), Dimensionality Reduction Based Methods (Green), Discretization Based Methods (Yellow), Graph Based Methods (Orange), Hybrid Based Methods (Pink)*

***Table 8:*** *The Included Methods for 3DSS Along with their Code Implementations*

| Algorithm | Category | Year | Code |
|---|---|---|---|
| PointNet | PB | 2017 | https://github.com/charlesq34/pointnet |
| PointNet++ | PB | 2017 | https://github.com/charlesq34/pointnet2 |
| SqueezeSeg | DRB | 2017 | https://github.com/BichenWuUCB/SqueezeSeg |
| OctNet | DB | 2017 | https://github.com/griegler/octnet |
| PointCNN | PB | 2018 | https://github.com/yangyanli/PointCNN |
| RSNets | PB | 2018 | --- |
| SqueezeSegV2 | DRB | 2018 | --- |
| TangenConv | DRB | 2018 | --- |
| SPLATNet | DB | 2018 | https://suhangpro.github.io/splatnet/ |
| SPGRaph | HB | 2018 | https://github.com/loicland/ superpoint_graph. |
| KPConv | PB | 2019 | https:// github.com/ HuguesTHOMAS/ KPConv |
| RangeNet++ | DRB | 2019 | https://github.com/PRBonn/lidar-bonnetal. |
| MinkowskiNet | DB | 2019 | https://github.com/StanfordVL/MinkowskiEngine |
| DGCNN | GB | 2019 | https://github.com/WangYueFt/dgcnn |
| PVCNN | HB | 2019 | https://github.com/mit-han-lab/pvcnn |
| RandLA-Net | PB | 2020 | https://github.com/QingyongHu/RandLA-Net<br>https://github.com/aRI0U/RandLA-Net-pytorch |
| SqueezeSegV3 | DRB | 2020 | https://github.com/chenfengxu714/SqueezeSegV3. |
| SalsaNext | DRB | 2020 | https://github.com/TiagoCortinhal/SalsaNext |
| MPF | DRB | 2020 | --- |
| PolarNet | HB | 2020 | https://github.com/edwardzhou130/PolarSeg |
| TORNADO-Net | HB | 2020 | --- |
| JS3C-Net | HB | 2020 | https://github.com/yanx27/JS3C-Net. |
| SPVConv | HB | 2020 | https://github.com/mit-han-lab/torchsparse<br>https://github.com/mit-han-lab/spvnas |
| LatticeNet | HB | 2020 | https://github.com/AIS-Bonn/lattice net. |
| KPRNet | HB | 2020 | https://github.com/DeyvidKochanov-TomTom/kprnet |
| SCF-Net | PB | 2021 | https://github.com/leofansq/SCF-Net |
| FG-Net | PB | 2021 | https://github.com/KangchengLiu/Feature-Geometric-Net-FG-Net |
| PTv1 | PB | 2021 | https://github.com/POSTECH-CVLab/point-transformer (Unofficial)<br>https://github.com/Pointcept/Pointcept |
| BAAF-Net | PB | 2021 | https://github. com/ShiQiu0419/BAAF-Net |
| FPS-Net | DRB | 2021 | https://github.com/xiaoaoran/FPS-Net |
| MINet | DRB | 2021 | https://github.com/sj-li/MINet |
| Efficient BEV | DRB | 2021 | --- |
| DeepGCN | GB | 2021 | https://github.com/lightaime/deep_gcns_torch<br>https://github.com/lightaime/deep_gcns |
| $(AF)^2$_S3Net | HB | 2021 | --- |
| Cylinder3D | HB | 2021 | https://github.com/xinge008/Cylinder3D |
| 3D-MiniNet | HB | 2021 | https://sites.google.com/a/unizar.es/semanticseg/ |
| PTv2 | PB | 2022 | https://github.com/Gofinge/PointTransformerV2<br>https://github.com/Pointcept/Pointcept |
| GFNet | DRB | 2022 | https://github.com/haibo-qiu/GFNet |
| 2DPASS | HB | 2022 | https://github.com/yanx27/2DPASS |
| DeepViewAgg | HB | 2022 | https://github.com/drprojects/DeepViewAgg |
| Meta-RangeSeg | HB | 2022 | https://github.com/songw-zju/Meta-RangeSeg |
| UniSeg | HB | 2023 | https://github.com/PJLab-ADG/PCSeg |
| PTv3 | HB | 2024 | https://github.com/Pointcept/PointTransformerV3 |

In section 2 an analysis on previous review papers is made to define a unified taxonomy scheme for the 3DSS methods. Regardless, the category in which the 3DSS algorithms belong to i.e., Point, Dimensionality Reduction, Discretization, Graph and Hybrid Based, they aim to handle similar issues, like the computational cost and memory consumption, the information loss during the algorithm execution, the properties of the 3D point cloud e.g., sparsity and irregularity and the introduction of a novel feature extraction strategy etc. However, based on their category, the advantages and disadvantages of them as long as the new research directions are unique.

**Regarding the 3DSS Method Category:** Specifically, the point based methods are applied directly on the given 3D point cloud, extracting meaningful 3DSS point features and thus there is no need of extra computational time to create an intermediate representation of the data. To this end, the 3D spatial structure information of the data is preserved without losing the detailed geometric information [125]. A crucial part on 3DSS using point based methods is to preserve the detailed geometric information of the data during the execution of the algorithm. Regarding this task, the sampling method used to define each neighborhood centroid point is one of the main parts. The most common sampling method is the Farthest Point Sampling (FPS) one [68], [69], [78], [138]. However, several approaches mentioned that the FPS is time-consuming [70], [139] while other sampling methods like Random Sampling (RS) could reduce the execution time as long as preserving high-end results. In fact, RS has also some drawbacks due to the random picking of points e.g., to dropout useful point features [70]. Hence, an investigation on new strategies on point sampling, to improve the efficiency and efficacy of the point based methods is recommended. In general, [70] presented an in depth analysis of many sampling methods and thus it could be used as a source.

Furthermore, the definition of the points neighborhood plays a significant role in the preservation of the local information of the point cloud as long as to the execution time of the algorithm. The most common approaches to define the neighborhood of points is the k Nearest Neighbors (kNN) and the Ball Query algorithms. In general, the Point Based methods spend most of the execution time to define the point's neighborhood i.e., to handle the random memory access or in other words to extract the point neighbors into the continuous space, rather than extracting features [95], [104], [113], [114]. Frequently, the point based methods are characterized as unreliable and time consuming especially for large scale point clouds due to the neighborhood definition step [1], [90], [91], [94], [116], [125], [139], [140], [141]. In fact, the point based methods are applied on large scale point clouds using point cloud partitioning approaches. However, the partitioning operation harms the global consistency of the point cloud resulting to weaker global features [142]. Furthermore, the kNN algorithm is not sufficient for modeling the local neighborhood of points [143]. Thus, efficient techniques on the definition of the points' neighborhood should be investigated in order to alleviate the issue of applying point based methods on large scale point clouds [144]. Finally, a more efficient definition of point's neighborhood will further improve the preservation of the local information encapsulated into the points' neighborhoods and thus to enable the application of the point based 3DSS methods in more detailed applications e.g. point cloud Serialization.

Moreover, the local feature aggregation method which is applied on the points neighborhood to transform the 3DSS local features, is crucial for the preservation of the local information. In fact, the extraction of the local information from the points' neighborhood is demanding due to the point cloud properties and the lack of explicit relationship among the 3D points [145]. Furthermore, the existing 3DSS approaches do not thoroughly capture the local information [51], [76], [146]. In general, the local feature aggregation operations of the point based 3DSS algorithms are commonly used in the maximum or summation operation. However, these operations tends to lose the information of the geometric structures [147]. Even more deeply, the way that the local aggregation is applied on each category affects the performance of the model. In general, the existing 3DSS algorithms process all the categories using the same aggregation operation which could result in confusion among similar categories [148], [149]. Additionally, the complex details presented into the point clouds lead to pattern imbalances either between the categories, resulting to the developed algorithms to learn only the dominant cases [150] or inside the same category resulting to reduce the models performance [151] . Overall, an improvement on the local aggregation operators will positively affect diversly the point based 3DSS algorithms and thus an investigation of such part is recommended. In general, [152] presented an in depth analysis of the local aggregation operators in which they surprisingly concluded that the sophisticated local aggregation operators performed similarly to the conventional ones using the same residual network. Hence, it could be used as a source for further exploration.

Furthermore, most of the 3DSS algorithms investigate the local feature extraction giving less attention to the global features, which are quite important for the performance of the 3DSS algorithms [153], [154], [155], [156]. Thus, an investigation of the global feature extraction process could be performed to improve the performance of the 3DSS algorithms. Regardless the local and global feature extraction, multiscale feature extraction is also investigated aiming to find multiresolution information and thus improving the performance of the 3DSS methods [157], [158]. Finally, point based methods require less GPU memory in comparison to the discretization based methods [113] while they achieved high-end results in many 3D tasks [90], [152].

In general, the Dimensionality Reduction Based (DRB) methods project the given point cloud into a lower dimension space, perform the semantic segmentation of the data in that space and then reproject it into the 3D space. Using the aforementioned methodology, the DRB methods try to alleviate the unstructured and irregularity properties of the 3D point clouds by constructing a regular representation of the data.

However, the creation of such representation costs extra computation time and also results to a loss of information e.g., the geometric information and the details of the point cloud, due to the projection of the 3D data to the lower dimension space [1], [70], [73], [78], [87], [97], [116], [118], [136], [139], [152], [159]. Although the extra computation time required for the creation of the new representation, the DRB methods overcome the expensive 3D computations presented in the point based and graph based methods [107]. Besides, the projection process alters and abandons the 3D topology and the geometric relations of the data [118], [125], [136], [160]. However, the DRB methods could be benefited from the mature 2DSS algorithms and 2DSS datasets and thus to follow their advancements [95], [97], [107]. Overall, the DRB methods seems to be fast, but without exploiting the benefits of the 3D data [91], [123].

Moreover, the DRB methods require post-processing steps to reproject the semantic information gathered into the lower space to the 3D space. In fact, the performance of the 3DSS is hampered due to the smoothed labels that are generated in the lower dimension space and reprojected into the 3D space [87], [122] and thus a refinement technique is commonly applied to achieve high end results. This problem was formulated as the label re-projection problem while the first approaches tried to alleviate it using Conditional Random Fields (CRFs) [82]. In general, the most common approach which handles the label re-projection problem is that presented in RangeNet++ using the kNN algorithm [87]. Recently, an alternative post processing solution, to avoid the kNN approach was presented exploiting the KPConv [73] network [2], [122]. In general, new post processing techniques to alleviate the label re-projection problem could be explored.

In fact, the DRB methods stack several channels to create multi-channel range images from the 3D point cloud data. Commonly, the created images contain the x, y, z, range and remission or intensity channels. However, the feature distribution of such images varies and are different from the RGB images. Xu *et al.* (2020) mentioned that the convolution operator performs poorly using spatially-varying feature images. Additionally, Xiao *et al.* (2021) stated that each channel has unique characteristics and thus should be treated differently. In general, an investigation of how the developed algorithm will fully exploit the information presented into the generated multi-channel images of the DRB methods, will be beneficial.

Moreover, the DRB methods most commonly use the spherical projection of the 3D data. However, there are other projection like the Bird's Eye View (BEV) which are exploited to perform 3DSS. Jiang *et al.* (2023) stated that the methods which exploit the BEV projection could achieve real-time inference on 3DSS. Recently, a combination of such projections was used to perform 3DSS mentioning that each one contains a complementary information to the other [2], [97]. Thus, a further exploration of projections in order to fully exploit the 3D information of the point clouds could be performed e.g., using scan unfolding [122].

In fact, the point based methods and the discretization based (DB) methods have some similarities regarding the feature extraction process i.e., both extract features using points neighborhood however, the neighborhood definition is different i.e., using kNN or Voxels. The DB methods transform the unstructured point cloud into a regular representation e.g., voxel grid, on which the well-known convolution operation could be applied. However, the creation of the regular grid is a time consuming operation due to the cubic growth of voxels in respect to the density of the given point cloud and the requirements of the voxel grid resolution [1], [2], [70], [72], [78], [80], [87], [88], [97], [102], [104], [111], [113], [114], [139], [152], [159], [162], [163]. Additionally, the quantization of the given point cloud results to a loss of information especially when points of different categories are included to the voxels or a large scale point cloud is used [1], [93], [105], [113], [114], [139], [152], [159], [164]. Mainly, the computation cost of the DB methods was due to the large amount of empty voxels, i.e., of the empty space, involved into the computations. Thus, sparse structures like octrees, kd-trees, and hash-maps [67], [73], [80], [88], [102], [104], [162] and postprocessing techniques like CRFs [67], [80] were introduced to alleviate the aforementioned problem. However, the convolution kernels that are exploited by the sparse voxel based methods are small and thus harming the networks performance [165]. Furthermore, the voxel-based methods have good memory locality i.e., preserving the spatial relationship due to the regular grid representation [113], [139]. To conclude, the DB methods are commonly applied separately but also in combination with other methods e.g., point based, forming hybrid architectures. A further investigation of hybrid architecture exploiting the advantages of the DB methods is recommended. Additionally, the high dimensional lattices could be exploited as a base for the exploration of the multi modal 3DSS.

In general, the Graph Based (GB) methods aim to create a new graph representation and then to extract 3D features exploiting the convolution operation. However, the creation of such representation is time consuming [139]. Additionally, the GB methods are commonly limited to very shallow models [106]. However, the graph representation of the data preserves the local geometry of the data while it can capture complex shapes. Moreover, the edges could be used to carry features on them [105]. Finally, the graph representation could be exploited in different than the 3D Euclidean space e.g., feature space, [105] which is advantageous for the creation of hybrid methods.

The four main categories of the 3DSS methods i.e., Point, Dimensionality Reduction, Discretization and Graph are commonly combined to form architectures which leverage the advantages of each category and complements the disadvantages of them. In general, the receptive field plays a significant role to the performance of both 2D and 3D architectures. However, in 3D space the shape apart from the size of the receptive field, is also important [104], [107]. Hence, several hybrid approaches which exploit discretization based methods, tried to define voxels which follow the distribution of the LiDAR points e.g., using polar grid [107], cylindrical voxel shape with different size depending on the distance of the 3D point [118] or crisscross shape voxels [104]. In fact, the LiDAR based methods dominates the research on 3DSS. Hence, the different shape voxels follow the acquisition of LiDAR point clouds. A further exploration of the voxel shape regarding other types is interesting.

Moreover, the discretization based and point based methods are commonly combined to leverage the efficient memory locality and the extraction of fine grained details, respectively [113], [114], [116], [118] using two branch feature extraction approaches. Different fusion strategies e.g., early or late fusion could be explored using features gathered using different approaches.

Additionally, the complementarity among different data modalities is also explored mainly using combination of images and 3D point clouds [119], [124]. The general idea is that the images contain better textural and contextual information than the 3D point clouds. To conclude, the complementarity among the different categories of the 3DSS methods should be further explored to advance the performance and efficiency of the 3DSS algorithms.

**Regarding the Datasets and the Data:** In general, most of the 3DSS algorithms are trained using supervised learning i.e., having available 3D ground truth information. In section 3 several 3DSS datasets for indoor and outdoor environments are presented. In fact, most of the datasets were created using an RGB-D camera and a LiDAR sensor in indoors and outdoors environments respectively. Additionally, even the available datasets use the same sensors like the Matterport camera (Indoor) and the Velodyne LiDAR (Outdoor). On the one hand, the combination of datasets is more straightforward due to the same data acquisition sensor type [166]. On the other hand, datasets with a different acquisition method e.g., using photogrammetric approaches or different sensors, should be released. To be more specific, the 3D point clouds created using a LiDAR sensor are strongly differ to those created using photogrammetry or terrestrial laser scanners in respect to point density and point distribution. Thus, more datasets using such techniques should be released to explore the 3DSS concept using them.

Additionally, there are not so many 3DSS datasets exploring multi-modality. Specifically, multi-modal 3DSS segmentation should be explored in regard of using complementary information using different types of data to improve the 3DSS performance e.g., using images in combination with 3D point clouds or other type of data like text. The aforementioned problem is also referred as sensory gap problem [9]. However, the aforementioned exploration should be assisted with well-constructed benchmark datasets which currently are in lack.

Furthermore, in order for the 3DSS algorithms to achieve better generalization, data from different geographic locations should be available. In fact, the 3DSS datasets are not span over different geographic locations [166]. Thus, datasets from different places on earth should be released to advance the interpretation of the 3DSS algorithms.

Moreover, a common problem regarding the datasets is the class imbalance problem [91], [93], [95]. More concretely, the number of points of each category significantly differ to each other e.g., the class motorcycle will have less points than the class street or pavement resulting to underperformance of the 3DSS on these classes. Additionally, the intra class geometry between the instances of each class plays a significant role on the performance of the 3DSS algorithms. A further exploration of the class-imbalance problem and the investigation of the differences in the intra class geometry, is recommended to improve the performance of the 3DSS algorithms on these categories.

In general, the available 3DSS datasets offer static information i.e., in specific temporal information. Hence, the creation of spatio temporal datasets of the same scene will be beneficial for several applications like constructions inspection etc.

Finally, applications oriented datasets such as in construction, cultural heritage, climate change etc. are important to be created, to advance the 3DSS performance in each application. However, this process is time-consuming and tedious while requires experts for the labeling annotation process. A counterpart of the creation of real application oriented datasets is the creation of application oriented synthetic datasets. However, there are limitations due to the domain shift problem, i.e., different acquisition circumstances between the real and the synthetic data.

**Regarding the application:** In general, the application plays a significant role to the characteristics of the using 3DSS algorithm. More detailed, some applications need real-time 3DSS algorithms e.g., in an autonomous driving scenario. Moreover, other applications need lightweight architectures e.g., using embedded systems while other applications require unique implementations due to the large scale of the available data e.g., urban scale 3DSS. Afterall, a further exploration of methods or the advancement of existing methods, regarding a specific application is recommended.

**Regarding the Learning Approach:** In fact, most of the available 3DSS methods use a supervised learning approach i.e., they exploit available 3D ground truth data to train the algorithms. However, the creation of such data is a time consuming and tedious process and is commonly referred as the data hungry problem [9] i.e., the existing algorithms require a large amount of fine annotated data to be trained on [52], [86], [167], [168]. Additionally, the creation of manual annotated data is error prone especially in 3D domain [169], [170]. To this end, other learning approaches like semi-supervised, weakly-supervised, few/zero shot learning or even unsupervised learning have been proposed to alleviate the aforementioned limitation [168], [171]. To be more specific, in 3D space the exploitation of these learning techniques is crucial regarding the difficulty to find application oriented datasets. Additionally, such techniques could be used to enrich the available datasets with new classes, unseen objects i.e., the open set problem. Moreover, weaker types of annotations have been released such as the scribble annotations [128], to define an easier way to create annotated datasets. Furthermore, continual learning techniques could be adopted to further improve the interpretation of the 3DSS algorithms. To this end, the exploration of extracting further knowledge about a scene using the available information could also be explored, commonly referred as the semantic gap problem [9]. Afterall, the exploration of different than the supervised learning techniques for the 3DSS is recommended, to alleviate the data hungry problem and to define open set 3DSS approaches.

## 8. CONCLUSIONS

In this effort, we surveyed in detail many of the most notable deep learning 3DSS algorithms and datasets of the previous years. Firstly, based on the analysis of the existing methods along with the investigation of previous review papers, we propose a unified taxonomy scheme. In detail, we categorize the 3DSS methods into the -Point, -Dimensionality Reduction, -Discretization, Graph and Hybrid Based methods Additionally, we present a thorough review of the existing indoor and outdoor 3DSS benchmarks along with their characteristics and evaluation metrics. Moreover, a not exhausting but informative presentation of the loss functions used in 3DSS is included. Furthermore, a fruitful discussion of the examined 3DSS algorithms and datasets, is presented to foster new research directions and applications in the field of 3DSS. Finally, a GitHub repository (https://github.com/thobet/Deep-Learning-on-3D-Semantic-Segmentation-a-Detailed-Review) which includes a quick classification of over 400 3DSS algorithms is included.

**ACKNOWLEDGMENT**
This manuscript was awarded the Laura Bassi Scholarship - Spring 2024, by the Bassi Foundation and Editing Press as part of the research of the first author towards his PhD at NTUA.